\documentclass{article}

\usepackage[nonatbib, preprint]{neurips_2022}

\usepackage[numbers,sort&compress]{natbib}
\usepackage[utf8]{inputenc} 
\usepackage[T1]{fontenc}    
\usepackage{hyperref}       
\usepackage{url}            
\usepackage{amsfonts}       
\usepackage{nicefrac}       
\usepackage{microtype}      
\usepackage{xcolor}         

\usepackage{graphicx}\graphicspath{{figures/}}
\usepackage{subfigure}
\usepackage{booktabs,enumitem} 
\usepackage{xspace}
\usepackage{algorithm}
\usepackage[noend]{algpseudocode}
\usepackage{varwidth}
\usepackage{wrapfig}

\usepackage{thmtools}
\declaretheorem[style=plain,numberwithin=section,name=Theorem]{theorem}
\declaretheorem[style=definition,sibling=theorem,name=Definition]{definition}

\usepackage{bm}
\usepackage{xparse} 
\NewDocumentCommand \V { O{} m } {{\bm{#1\mathbf{\MakeLowercase{#2}}}}}

\newcommand{\phaseone}{pre-training phase\xspace}

\newcommand{\phasetwo}{mask search phase\xspace}

\newcommand{\phasethree}{sparse training phase\xspace}

\newcommand{\goodinit}{matching initialization\xspace}

\newcommand{\rt}{t_r}

\newcommand\optparen[1]{\ifthenelse{\equal{#1}{}}{}{(#1)}}

\newcommand{\Reals}{\mathbb{R}}

\usepackage[hide=true,marginparwidth=1in]{marginalia}

\usepackage[capitalize,noabbrev]{cleveref}

\title{Lottery Tickets on a Data Diet:
Finding Initializations with Sparse Trainable Networks}

\author{Mansheej Paul$^1$\thanks{Equal contribution. Correspondence to:  \texttt{\{mansheej,bwlarsen\}@stanford.edu; gkdz@google.com}}\ \ \ \ \ \ \ Brett W. Larsen$^{1*}$\ \ \ \ \ \ \   \\
\textbf{Surya Ganguli}$^{1,2}$\  \ \ \ \ \ \   \textbf{Jonathan Frankle}$^{3,4,5}$ \ \ \ \ \ \ \  \textbf{Gintare Karolina Dziugaite}$^{6,7}$\\
$^1$Stanford \ \ \ \  $^2$Meta AI \ \ \ \ $^3$MIT \ \ \ \ $^4$MosaicML  \ \ \ \ $^5$Harvard \ \ \ \ $^6$Google Brain \ \ \ \ $^7$Mila; McGill}

\begin{document}

\maketitle

\begin{abstract}
A striking observation about iterative magnitude pruning (IMP; \citealt{frankle2019linear}) is that---after just a few hundred steps of dense training---the method can find a sparse sub-network that can be trained to the same accuracy as the dense network. 
However, the same does not hold at step 0, i.e., random initialization. 
In this work, we seek to understand how this early phase of pre-training leads to a good initialization for IMP both through the lens of the data distribution and the loss landscape geometry.
Empirically we observe that, holding the number of pre-training iterations constant, training on a small fraction of (randomly chosen)  data suffices to obtain an equally good initialization for IMP.
We additionally observe that by pre-training only on ``easy'' training data we can decrease the number of steps necessary to find a good initialization for IMP compared to training on the full dataset or a randomly chosen subset.
Finally, we identify novel properties of the loss landscape of dense networks that are predictive of IMP performance, showing in particular that more examples being linearly mode connected in the dense network correlates well with good initializations for IMP.
Combined, these results provide new insight into the role played by the early phase training in IMP.
\end{abstract}
\newcommand{\terr}[1]{\mathrm{err}(#1)}

\section{Introduction}
\label{sec:intro}
\begin{figure}
  \begin{minipage}[c]{0.5\textwidth}
  \includegraphics[width=\linewidth]{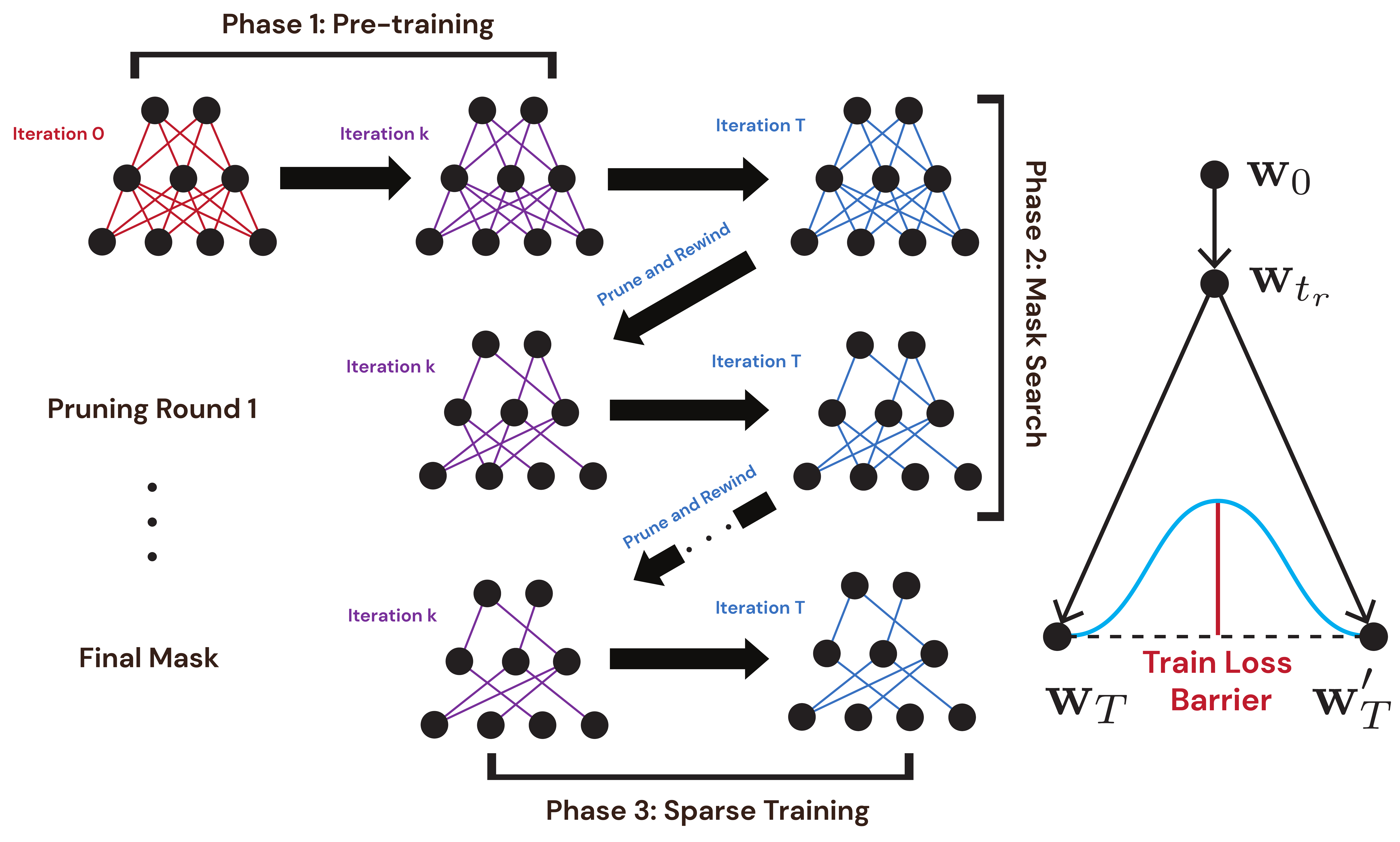}
  \end{minipage}\hfill
  \begin{minipage}[c]{0.49\textwidth}
    \caption{\textbf{Left:} Three phases of iterative magnitude pruning (IMP) with weight rewinding \citep{frankle2019linear}. A dense network is trained in the \phaseone for $\rt$ iterations, where $\rt$ is referred to as the rewinding iteration, and $\V{w}_{\rt}$ is the parameters of the network at the rewinding point. The \phasetwo produces a sparse subnetwork at a desired sparsity level by iteratively training, pruning the smallest magnitude weights, and rewinding to $\V{w}_{\rt}$. The \phasethree trains the final sparse subnetwork to convergence, starting with weights $\V{w}_{\rt}$. \textbf{Right:} Illustration of computing the train loss barrier for initialization $\V{w}_{t_1}$.  $\V{w}_T$ and $\V{w}'_T$ are trained with different data order.}
\label{fig:IMP-Illustration}
  \end{minipage}
\end{figure}

Modern deep neural networks are often trained in the massively over-parameterized regime. 
Though these networks can eventually be pruned, quantized, or distilled into smaller networks, the resources required for the initial over-parameterized training poses a challenge to the democratization and sustainability of AI. 
This raises a fundamental question: under what circumstances can we efficiently train sparse networks?
Recent work on the lottery ticket hypothesis \citep{frankle2018lottery,frankle2019linear} has shown that, after just a few hundred steps of pre-training, a dense network contains a sparse sub-network that can be trained without any loss in performance.
Finding this sparse sub-network currently requires multiple rounds of training to convergence, pruning, and rewinding to the pre-train point, a procedure termed iterative magnitude pruning (IMP, \cref{fig:IMP-Illustration}; \citep{frankle2018lottery,frankle2019linear}),
Remarkably, even after all these rounds of training, we do not find trainable sparse sub-networks if we rewind to the random initialization; the first few hundred steps of dense network training is essential for finding sparse networks through IMP.
In this work, we seek to understand this \emph{very short but critical phase} of pre-training. 
In particular, we investigate the effect of training data and number of steps used during pre-training on the accuracy achieved by IMP.
We also explain how certain properties of the loss landscape allow us to predict whether we will find a \emph{\goodinit} (i.e.,  an initialization from which IMP ``succeeds'' in finding a subnetwork that can match the accuracy of the unpruned network, to be formalized later). 

IMP proceeds in \emph{three} phases: 
an initial \emph{\phaseone} where (1) the dense weights are trained for a few hundred steps, followed
by (2) the resource-intensive \emph{\phasetwo}, during which we iteratively find a  sparse mask by training the network and rewinding the weights repeatedly,
finally, ending with
(3) the \emph{\phasethree}, when the sparse masked network is trained.
The weights learned after \phaseone---which serve as the initialization for \phasetwo---are referred to as the \emph{pre-trained initialization} (see \cref{fig:IMP-Illustration}). 

In this work, we focus on studying the \phaseone.
To probe sufficient information needed to arrive at a \goodinit for IMP, we modify the pre-training phase by training with different pruned datasets and varying the number of pre-training steps; we then compare the performance of the resulting sub-networks across a range of sparsities.
Following \citet{paul2021deep}, we prune the datasets both randomly and according to EL2N scores, which estimate example difficulty by measuring early training performance in an ensemble of networks.
We next turn to characterizing the loss landscape properties of \goodinit for IMP.
\citet{frankle2019linear} investigate the relationship between successful initializations for IMP and linear mode connectivity on the \emph{sparse} initialization for the entire dataset; here, we investigate this relationship on the \emph{dense} initialization on a per-example basis.
This enables us to identify signatures of the \emph{dense} loss landscape correlated with a \goodinit.
Thinking more broadly, we explore the relationship between sparse trainability and training stability. 
\citet{gilmer2021loss} demonstrate that, in hyperparameter regimes in which early training is unstable, learning rate warmup helps stabilize training.
We build on this by investigating if the same pruned datasets which decreased the amount of pre-training required to find a \goodinit for IMP also decrease the required duration of learning rate warmup.

Overall, this empirical analysis provides new insight into the important and mysterious role that pre-training the dense network plays in IMP.
Intriguingly, finding a matching initialization for IMP is an example of a problem in which the dominant paradigm of ``more data, more training'' is not optimal. 
Indeed, by carefully choosing the training examples, we can not only use a very small subset of the data but also reduce the training time required to solve this problem. 
From a scientific perspective, this suggests that different phases of training play different roles in the optimization process. Identifying and characterizing these phases will not only lead to a deeper understanding of the optimization of deep neural networks but also allow us to design training strategies that are optimal for each phase.  From the practical standpoint, data loading is often an expensive part of the training process and can be a bottleneck especially early in training.  By identifying the essential subset for this phase of training, our work may enable strategies to eliminate this bottleneck.
\vspace{-1em}

\paragraph{Contributions.}

\newcommand{\finding}[1]{\textbf{#1}}
We find empirical evidence for the following statements:
\vspace{-1em}
\begin{itemize}[leftmargin=1em]
	\item \finding{In the \phaseone, only a small fraction of the data is required to find a \goodinit for IMP:} On standard benchmarks, across all sparsity levels we evaluated,
	we find that we can match accuracy by training on a small fraction of all of the available training data, selected randomly. (As we vary the amount of training data in \phaseone, the number of training iterations is held fixed.)  
	Note that this observation changes if random label noise is introduced, in which case it becomes important to select easy (small EL2N score) examples.
	\item \finding{The length of the \phaseone can be reduced if we train only on the easiest examples:} Informally, training on a small subset of ``easy-to-learn'' training examples produces a better rewinding point than training on all data for the same number of iterations;
  \item \finding{The quality of a pre-trained initialization for IMP correlates with more examples being linearly mode connected in the dense network:} 
	This result complements the empirical evidence produced by \citet{frankle2019linear} connecting linear mode connectivity and the performance of IMP. 
  \item \finding{Training on easy data, which reduces the amount of pre-training required to find a \goodinit for IMP, does not reduce the amount of training required to stablize the network via learning rate warmup:} Running learning rate warmup on easy data cannot replicate the warmup effect on the top loss Hessian eigenvalue (\citep{gilmer2021loss}), but a small fraction of medium-hard to hard data suffices.
\end{itemize}

\section{Background, Methods, and Related Work}
\label{sec:methods}

We consider standard neural network training on image classification.
Let $S = \{(\V{x}_n,\V{y}_n)\}_{n=1}^N$ denote training data, 
let $\V{w}_t \in \Reals^D$ denote model parameters (weights) of the neural network,
and let $\V{w}_1, \V{w}_2, \dots$ be the iterates of (some variant) of SGD, 
minimizing the training \emph{loss}, i.e., average cross-entropy loss over the training data. 
For a given training example $\V{x}_n$, let $f(\V{w}, \V{x}) \in \Reals^K$ denote the logit outputs of the network for weights $\V{w}$ and $p(\V{w}, \V{x}) = \sigma(f(\V{w}, \V{x}))$ be the probability vector returned by passing the logits through the softmax operation $\sigma$.
By the \emph{loss (error) landscape}, we mean the training loss (error), viewed as a function of the parameters.
By training and test \emph{error}, we mean the average 0-1 classification loss.

\textbf{Lottery ticket subnetworks.}
The \emph{lottery ticket hypothesis} \citep{frankle2018lottery} states that any standard neural network ``contains [at initialization] a subnetwork that is initialized such that---when trained in isolation---it can match the test accuracy of the original network after training for at most the same number of iterations.''
Although such \emph{matching subnetworks} (those that can train to completion on their own and reach full accuracy by following the same procedure as the unpruned network) are not known to exist in general at random initialization, they have been shown to exist after \emph{pre-training} the dense network for a short amount of time (Phase 1 in Figure \ref{fig:IMP-Illustration}) before pruning \citep{frankle2019linear, yu2020playing, chen2020lottery, vischer2021lottery, kalibhat2020winning}.

Empirical evidence for this phenomenon comes via a procedure that finds such subnetworks retroactively after training the entire network.
This procedure, called \emph{Iterative Magnitude Pruning} \citep[IMP; ][]{frankle2019linear} is based on standard iterative pruning procedures \citep{han2015learning}, and can be decomposed into three phases (Figure \ref{fig:IMP-Illustration}), outlined in \cref{alg:imp}.

\begin{algorithm}[h!]
    \small
    \caption{IMP rewinding to step $\rt$ and $N$ iterations.}
    \begin{algorithmic}[1]
    \State Create a network with randomly initialization $\V{w}_0 \in \mathbb{R}^d$.
    \State Initialize pruning mask to $\V{m} = 1^{d}$.
    \State Train $\V{w}_0$ for $\rt$ steps to $\V{w}_{\rt}$ \Comment{Phase 1: Pre-Training}
    \For{$n \in \{1, \ldots, N\}$} \Comment{Phase 2: Mask Search}
    \State Train the pruned network $\V{m} \odot \V{w}_{\rt}$ to completion. ($\odot$ is the element-wise product)
    \State \begin{varwidth}[t]{\linewidth}Prune the lowest magnitude 20\% of weights after training.\\Let $\V{m}[i] = 0$ if the corresponding weight is pruned.\end{varwidth}
    \EndFor
    \State Train the final network $\V{m} \odot \V{w}_{\rt}$. Measure its accuracy. \Comment{Phase 3: Sparse Training}
    \end{algorithmic}
    \label{alg:imp}
\end{algorithm}

This procedure reveals the accuracy of pre-training the dense network for $\rt$ iterations, pruning, and training the pruned network thereafter.
Phase 2 can be understood as an (expensive) oracle for choosing weights to prune at $\rt$.
Although IMP is too expensive to use as a practical way to speed up training, it provides a window into a possible minimal number of parameters and operations necessary to successfully train a network to completion in practice.
In our work, we extend this line of thinking, pursuing the minimal amount of data necessary to find and train these subnetworks.
This is especially tantalizing due to the potential positive interactions between sparsity and minimizing the data necessary for training.
\emph{The result is a deeper inquiry into the minimal recipe for successful training and, thereby, into the fundamental nature of neural network learning in practice.}

In this respect, the closest work to ours is an experiment in a larger compendium by \citet{Frankle2020The} showing that the standard pre-training phase could be replaced by a much longer self-supervised phase.
Like our experiments in Section \ref{sec:phase1}, that work aims to study what makes a \goodinit.
Our approach and findings are substantially different, however: we reduce the number of examples rather than changing the labels, and we show that---not only are a small set of examples (starting at $\approx 2\%$) sufficient for pre-training---but also that they make it possible to pre-train in \emph{fewer} steps.

There are many other ways to obtain pruned neural networks \citep[e.g.,][]{janowsky1989pruning, lecun1990optimal, han2015learning, zhu2017prune, evci2020rigging}.
The distinctive aspect of work on the lottery ticket hypothesis (and the one that makes it the right starting point for our inquiry) is that its goal is to uncover a minimal path from initialization to a trained network, regardless of the cost of doing so.
The aforementioned procedures target real-world efficiency for training and/or inference.

\textbf{Linear Mode Connectivity and the Loss Barrier.} We investigate the error landscape using the parent--child methodology and instability analysis of \citet{frankle2019linear} and \citet{fort2020deep}.
Writing $\terr{\V{w}}$ for the test error at the weights $\V{w}$,
the (test) loss barrier between two networks $\V{w}$ and $\V{w}'$ is
$\sup_{\alpha\in [0,1]}  [ \terr{\alpha\, \V{w} + (1-\alpha) \V{w}'} - (\alpha\, \terr{\V{w}} + (1-\alpha) \terr{\V{w}'}) ]$.
This quantity measures the maximum increase in loss above the average loss along the linear path connecting the two networks on the loss landscape.

The loss barrier at iteration $t$ (with parent weights $\V{w}_t$) is
the loss barrier between the (children) weights $\V{w}_T$ and $\V{w}'_T$, where $\V{w}_T$ and $\V{w}'_T$ are copies of $\V{w}_t$ trained to completion with the same procedure but different random seeds (minibatch order, GPU noise, data augmentation, etc.). See Figure \ref{fig:IMP-Illustration} right for a visualization. 
Empirically, the loss barrier is achieved near $\alpha = \frac 1 2 $, and so we compute it this way. 
The \emph{onset of linear mode connectivity (LMC)} is defined to be the iteration $\bar{t}$ such that, for all $t > \bar{t}$, the error barrier at $t$ is zero. 
We follow \citet{draxler2018essentially}, \citet{garipov2018loss}, and \citet{frankle2019linear} in considering the 0-1 loss barrier to be zero if it is less than 2\%.
Similar to \citet{paul2021deep}, we also measure the error (0-1 loss) or cross-entropy loss barrier on individual training examples to explore the loss landscape corresponding to different subpopulations.

\textbf{Ranking training examples.}
We define \emph{``easy/hard data''} as the data that is ranked low/high, respectively, by the EL2N score introduced by \citet{paul2021deep}. EL2N scores depend on the margin early in training, and, loosely speaking, higher average margin early in training means lower importance for generalization of the final trained model. 
This connection to margin suggests that easy data is learned first (has higher margin early in training, maintained throughout the rest of training). 
EL2N scores were derived from the size of the loss gradient, and are thus are highly correlated with the magnitude of the gradient. 
\vspace{-.5em}
\begin{definition}[EL2N Score] The EL2N score of a training sample $(\V{x}, \V{y})$ at iteration $t$ is defined as $\mathbb{E} \| p(\V{w}_t, \V{x}) - \V{y} \|_2$, where the expectation is taken over $w_t$ conditioned on the training data.
\end{definition}
\vspace{-.5em}
In our experiments (\cref{sec:phase1}), we vary the data that is accessible in the \phaseone of IMP defined above. 
We either choose the data that we feed to the algorithm at random while preserving class balance, or based on the EL2N scores.

\textbf{Stabilizing early training via learning rate warmup.} Learning rate warmup period can be seen as a form of pre-training, allowing one to eventually train at higher learning rates and larger batch sizes.
Training with large batches and high learning rates is desirable in practice, as it allows for more efficient GPU utilization and may reduce the total number of updates needed to achieve the desired accuracy. 
\citet{gilmer2021loss} empirically observe that learning rate warmup essentially improves the optimization by allowing the initial optimization trajectory to navigate to ``flatter'' optimization landscape, i.e., one with smaller highest loss Hessian eigenvalue. 

\begin{figure*}[!ht]
  \centering
	\includegraphics[width=\linewidth]{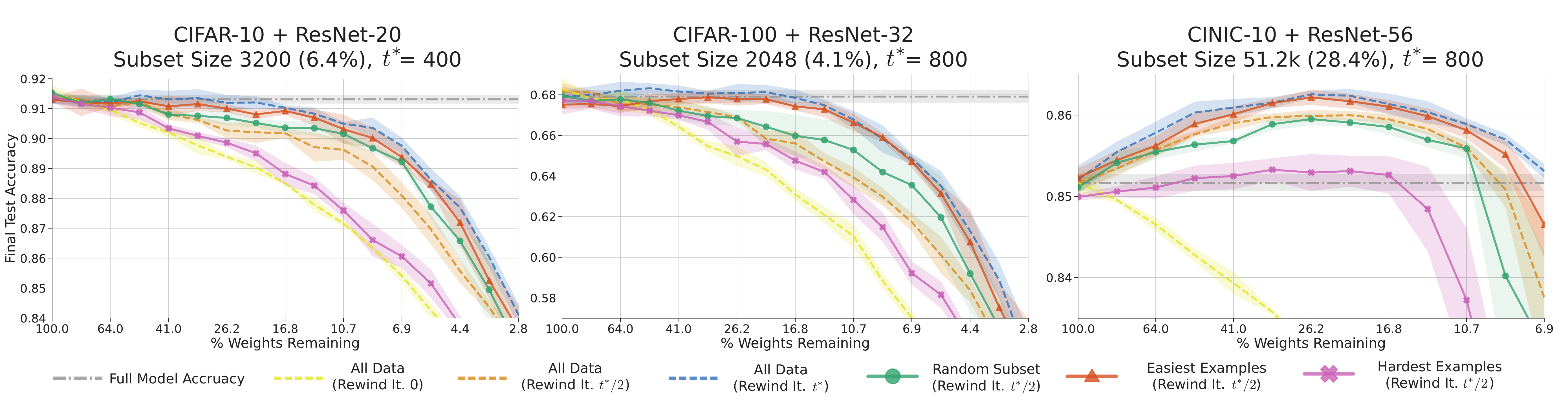}
	\vspace{-0.4cm}
	\caption{For a given rewind step $t_r = t^*/2$, training on a small fraction of random data during the \phaseone of IMP leads to matching initializations (compare the solid green with circles and dashed orange curves) across dataset, network, and hyperparameter configurations. Using just the easiest training examples during this phase produces a matching initialization for rewind point $t^*$ in just $t^*/2$ steps (compare the solid red with triangles and dashed blue curves).  Pre-training on the hardest examples is detrimental to the performance of the initialialization (solid pink curve with crosses). IMP with rewinding to initialization (dashed yellow curve) and the dense model (dashed grey curve) are used as baselines. For each dataset + network configuration, we present the best performing easy data subset size. For a sweep across subset sizes, see \cref{fig:Phase1Subset} and the \cref{sec:FullResults}.
    }
	\label{fig:Phase1AllSparsities}
\end{figure*}

\begin{figure}[!t]
	\centering
	\includegraphics[width=\linewidth]{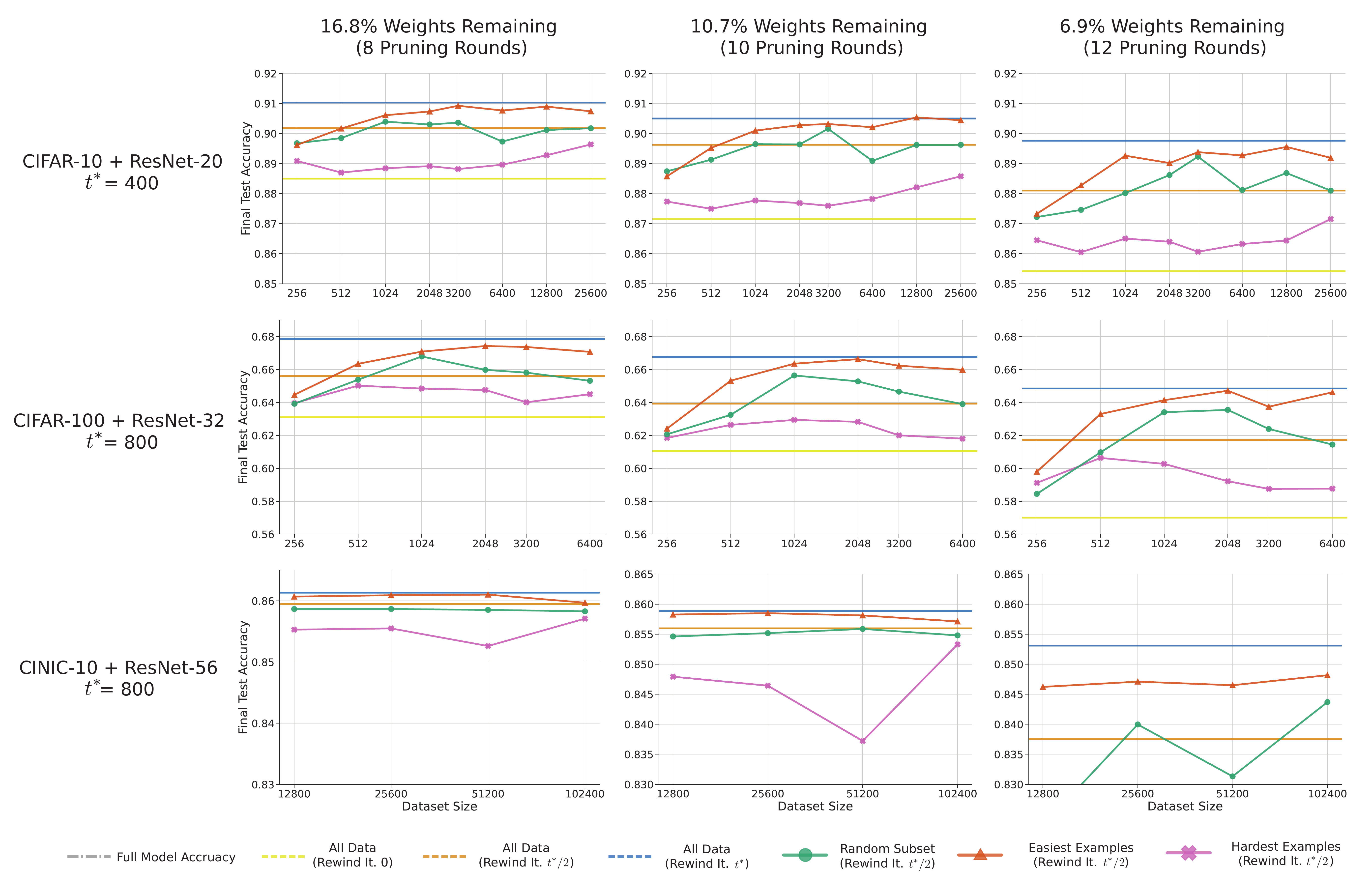}
	\vspace{-0.6cm}
	\caption{A summary of the the dependence on subset size for the style of experiments described in \cref{fig:Phase1AllSparsities}.  The first column represents the performance across subset size for the fixed sparsity 16.8\% weights remaining or 8 rounds of pruning.  The subsequent columns show the same for 10.7\% (10 pruning rounds) and 6.9\% (12 pruning rounds) respectively.  The horizontal lines correspond to baseline runs at rewind steps 0, $t^*/2$, and $t^*$ using all the data.  For CINIC-10 (bottom row), rewind step 0 and the hardest data subsets are not visible in some cases because their accuracies fall below the range displayed.
	}
	\label{fig:Phase1Subset}
\end{figure}

\vspace{-0.5em}

\section{The Role of Training Data Selection in Pre-Training}
\label{sec:phase1}

\vspace{-0.5em}

As can be seen in \cref{fig:Phase1AllSparsities}, when training sparse networks using IMP with rewind step $t_r = 0$, the final test accuracy of the sparse networks falls off rapidly with increasing sparsity. 
However, as we increase the rewind step $t_r$, the network performance improves across all sparsity levels and at a rewind step $t^*$, the network performs as well as or better than the dense network at high sparsities.
Informally, training the dense network for $t^*$ steps creates a matching initialization for IMP.
But what does the network learn in these first $t^*$ steps?
In this section, we take the first step towards answering this question by investigating which subsets of the training data are sufficient for finding a matching initialization. 
In order to compare networks trained on different subsets of data for different numbers of iterations, we introduce the notion of a \textit{matching initialization} with the following definitions.

\begin{definition}
Let $\V{w}^S_{t}$ be the dense network weights after training on a subset of the training data, $S$ until rewind step $t$.
Then for two data subsets $\{S, S'\}$, rewind times $\{t, t'\}$ and a given range of sparsities, $\V{w}^{S'}_{t'}$ is said to \emph{dominate} (weakly dominate) $\V{w}^S_{t}$ if sparse networks obtained from IMP with $\V{w}^{S'}_{t'}$ as the initialization achieve better (no-worse) accuracy than those obtained from IMP with $\V{w}^S_{t}$ as the initialization. 
\end{definition}

For a network trained on the full dataset for $t$ steps, we write $\V{w}_t$. In \cref{fig:Phase1AllSparsities}, we see that $\V{w}_{t^*}$ dominates $\V{w}_{t^*/2}$ which in turn dominates $w_{0}$. We investigate which data subsets $S$ and rewind steps $t$ lead to networks $\V{w}^S_t$ that dominate $\V{w}_{t^*}$ and $\V{w}_{t^*/2}$---such networks are called matching initializations.

\begin{definition}
A dense network $\V{w}^S_{t}$ is a matching initialization for rewind time $t^*$ if $\V{w}^S_{t}$ weakly dominates $\V{w}_{t^*}$. \footnote{Note that due to the stochasticity in training and risk measurements, we ignore small deviations in the final test accuracy.
Thus one rewinding point could weakly dominate another one even if at some sparsity levels their mean performance ``crosses'' over while approximately remaining within the standard error of one another.}
\end{definition}

\vspace{-0.5em}
We empirically find that certain surprisingly small subsets $S$ and rewind step $t_r<t^*$ lead to matchining initializations fore rewind time $t^*$.

\textbf{Experimental design.}
To evaluate the effect of the training subset size and composition on the quality of the pre-trained initialization, we train ResNet-20/ResNet-32/ResNet-56 on subsets of CIFAR-10/CIFAR-100/CINIC-10, respectively. 
The subset size $M$ is varied and subsets are chosen as follows:
(i) $M$ randomly selected examples, distributed equally among all classes; 
(ii) the easiest $M$ examples; 
(iii) the hardest $M$ examples. The easiest examples are those with the smallest EL2N scores and the hardest are the examples with the largest EL2N scores \citep{paul2021deep}.

Due to the significant computational demands of performing IMP with multiple pre-training schemes and replicates, we focus on a targeted set of pre-training iterations $\rt$.
In particular, we study the pre-training iteration $\rt = t^*$ where training on all examples leads IMP to find spare sub-networks that perform as well as the dense network for a large range of sparsities ($t^*$ = 400 for CIFAR-10 and 800 for CIFAR-100 and CINIC-10).
We also study the more challenging pre-training iteration of $\rt = \frac{t^*}{2}$, where pre-training on all data does not yield a \goodinit.
 When $M$ examples are not enough to train for $\rt$ iterations without replacement, we make multiple passes over the $M$ examples as necessary.  \Cref{fig:Phase1AllSparsities} shows the best performing easy data subset for each dataset across the full range of sparsities; \cref{fig:Phase1Subset} shows the performance across subset size at three fixed sparsities.
 
 \begin{figure}[!t]
	\centering
	\includegraphics[width=0.95\linewidth]{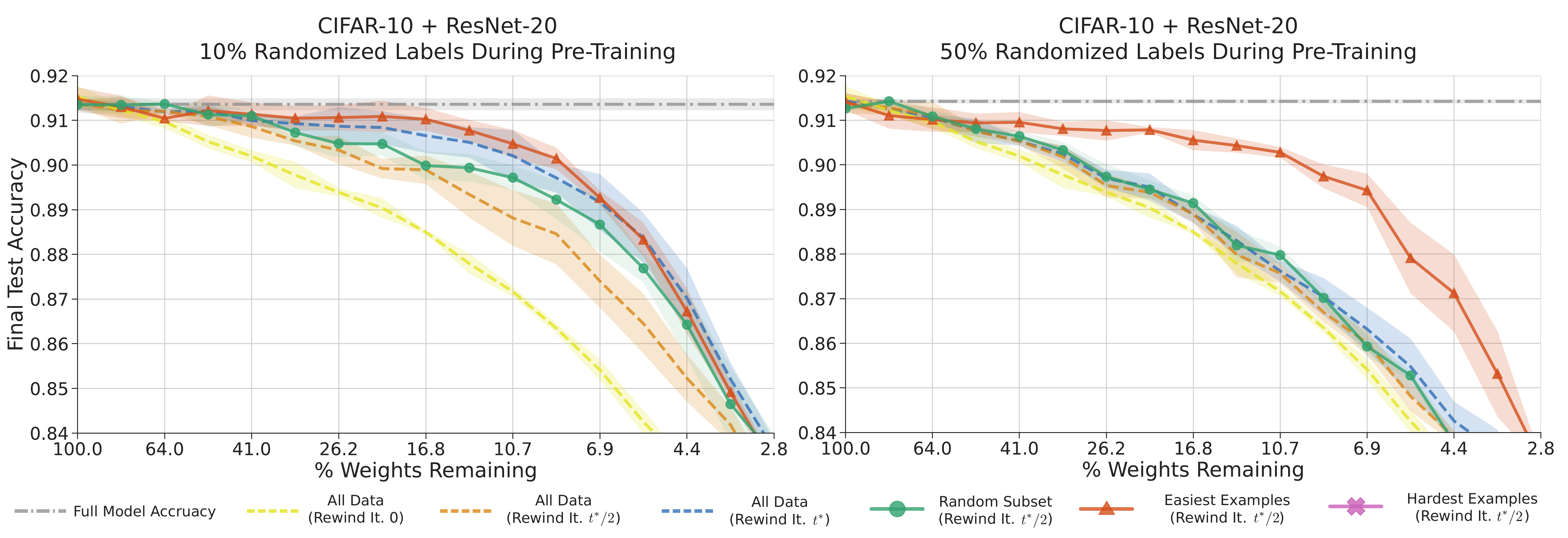}
	\vspace{-0.3cm}
	\caption{
	Pre-training on a random subset or all data is not robust to label noise during this initial phase of IMP.  However, pre-training with the easiest data as scored by EL2N scores computed from the corrupted dataset is robust.
	In both the left (10\% randomized labels) and right (50\% randomized labels), pre-training on the easiest data for $t^*/2$ iterations dominates all other pre-training schemes, including training on all data for $t^*$ iterations.
	Results for additional subset sizes are included in \cref{sec:FullResults}.}
	\label{fig:randomdatapretraining}
\end{figure}

\textbf{Randomly chosen examples.}
Pre-training the dense networks on small, randomly chosen subsets $S$ can lead to initializations for IMP, $\V{w}^S_{t_r}$, that dominate initializations $\V{w}_{t_r}$, trained on the entire training set for the same number of steps. 
In \cref{fig:Phase1AllSparsities} we see that for all dataset + network combinations, pre-training the dense network on a small random subset (solid green curve with circles; sizes ranging from 4.1\% for CIFAR-100 to 28.4\% for CINIC-10) for $t_r = t^*/2$ leads to initializations that (weakly) dominate those that were obtained from training the network for the same number of steps on all the data. This observation leads to a surprising suggestion: in these experiments, the subset size is smaller than the total number of images seen during the \phaseone; for the particular goal of finding a matching initialization of IMP, multiple passes through the same small dataset can be more beneficial than seeing more random data.

\textbf{Easiest examples (lowest EL2N scores).}
By pre-training on just the easiest examples (identified by lowest EL2N scores, solid red curve with triangles in \cref{fig:Phase1AllSparsities}), we can obtain matching initializations in fewer steps compared to training on the full dataset. 
In \cref{fig:Phase1AllSparsities}, we see that for all three dataset and network combinations and for the subset sizes shown, the initialization obtained from training on the easiest examples for $t_r={t^*}/2$ steps leads to matching initializations for $t^*$.

\textbf{Hardest examples (highest EL2N scores).}
Conversely, pre-training on the hardest examples (solid pink curve with crosses in \cref{fig:Phase1AllSparsities}) yields worse accuracies than pre-training on all examples or a random subset.  In fact, on CIFAR-10 the hardest examples perform little better than using no pre-training at all.
Interestingly, when training a dense network, these hardest examples are crucial for obtaining a network with good generalization properties \citep{paul2021deep}.
This suggests that while the hard example may be key later in training, repeated passes through easier examples should be the focus during the very early stages of training to quickly find a good initialization for IMP.

\textbf{Randomized labels during pre-training.}
Pre-training on a all data or a random subset is not robust to corruption with random label noise \citep{Zhang2016UnderstandingDL} during the pre-training phase.
As seen in \cref{fig:randomdatapretraining}, the higher the percentage of randomized labels, the lower the performance of these data subsets, and in particular, the pre-trained rewinding point becomes no better than a random initialization when $50\%$ of the labels are randomized. 
On the other hand, training on easiest data with EL2N scores computed on the corrupted dataset is robust to this noise (solid red curve with triangles in \cref{fig:randomdatapretraining}).
This is because examples with randomized labels are hard (\citep{paul2021deep}) according to this metric, and thus the easiest examples will select a subset of largely uncorrupted data. 

\textbf{Summary.}
Taken together, our results suggest that, finding a matching initialization for IMP at rewinding step $t^*$ is an interesting problem in which ``more data, more training'' is \textit{not} optimal; it is neither necessary to train on all the data nor to train for the full $t^*$ steps. 
In fact, we can get away with training on a surprisingly small dataset for as little as half the number of steps if we make multiple passes through the \textit{right} examples, in this case the easiest examples as defined by the lowest EL2N scores. 
In \cref{app:gradsize}, we show that  pre-training on easy data increases the norm of the minibatch gradient, refuting the hypothesis that pre-training on easy data effectively performs gradient clipping.
See \cref{app:otherhypothesis} for additional results on hypotheses for the role of easy data.

\vspace{-1em}

\begin{figure*}[!ht]
    \centering
    \subfigure[CIFAR-100, ResNet-32, $t^* = 800$ and $\rt=t^*/2=400$.]{
	\includegraphics[width=0.87\linewidth]{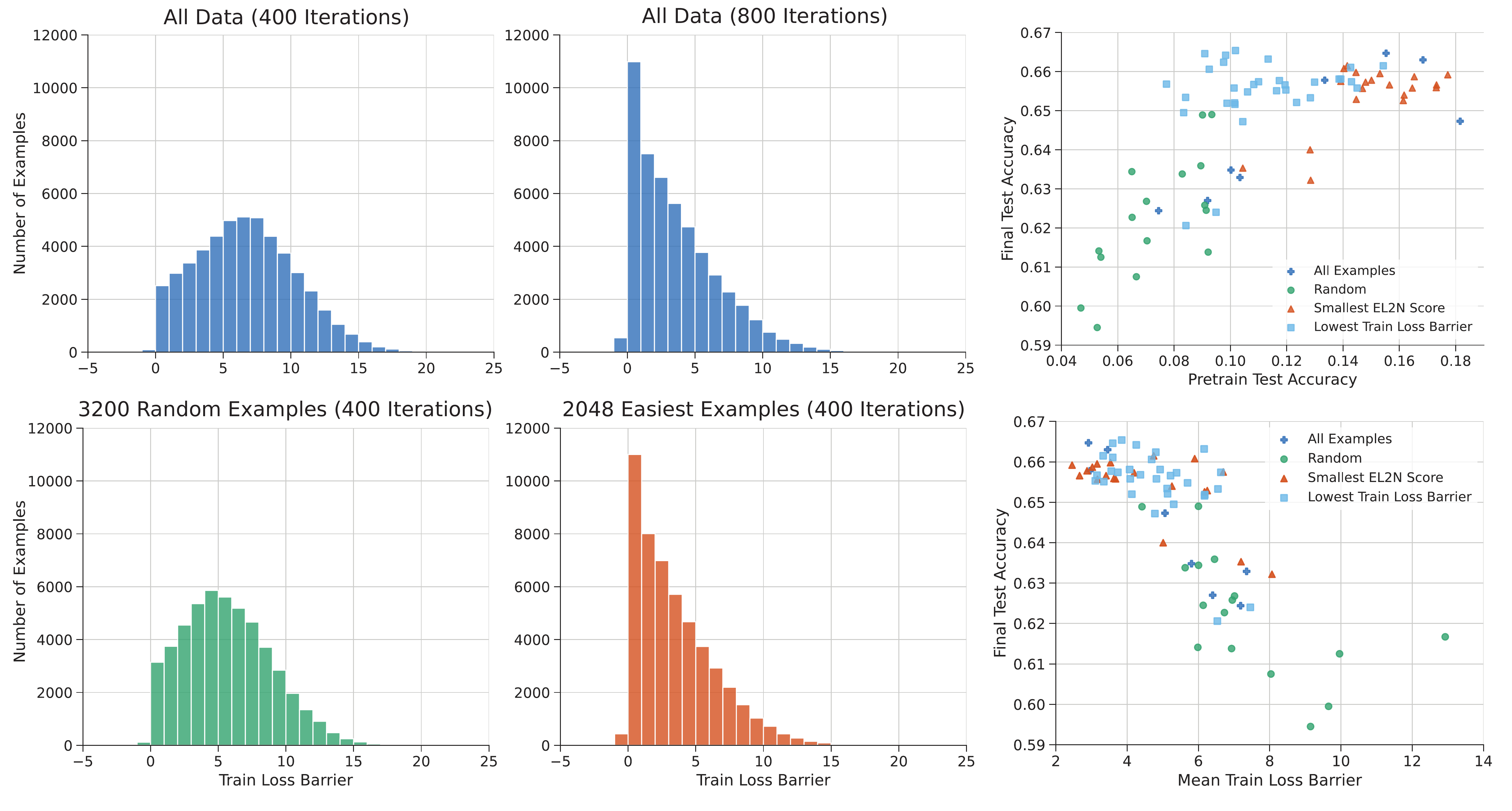}
	}
	\subfigure[CINIC-10, ResNet-56, $t^* = 400$ and $\rt=t^*/2=200$.]{
	\includegraphics[width=0.87\linewidth]{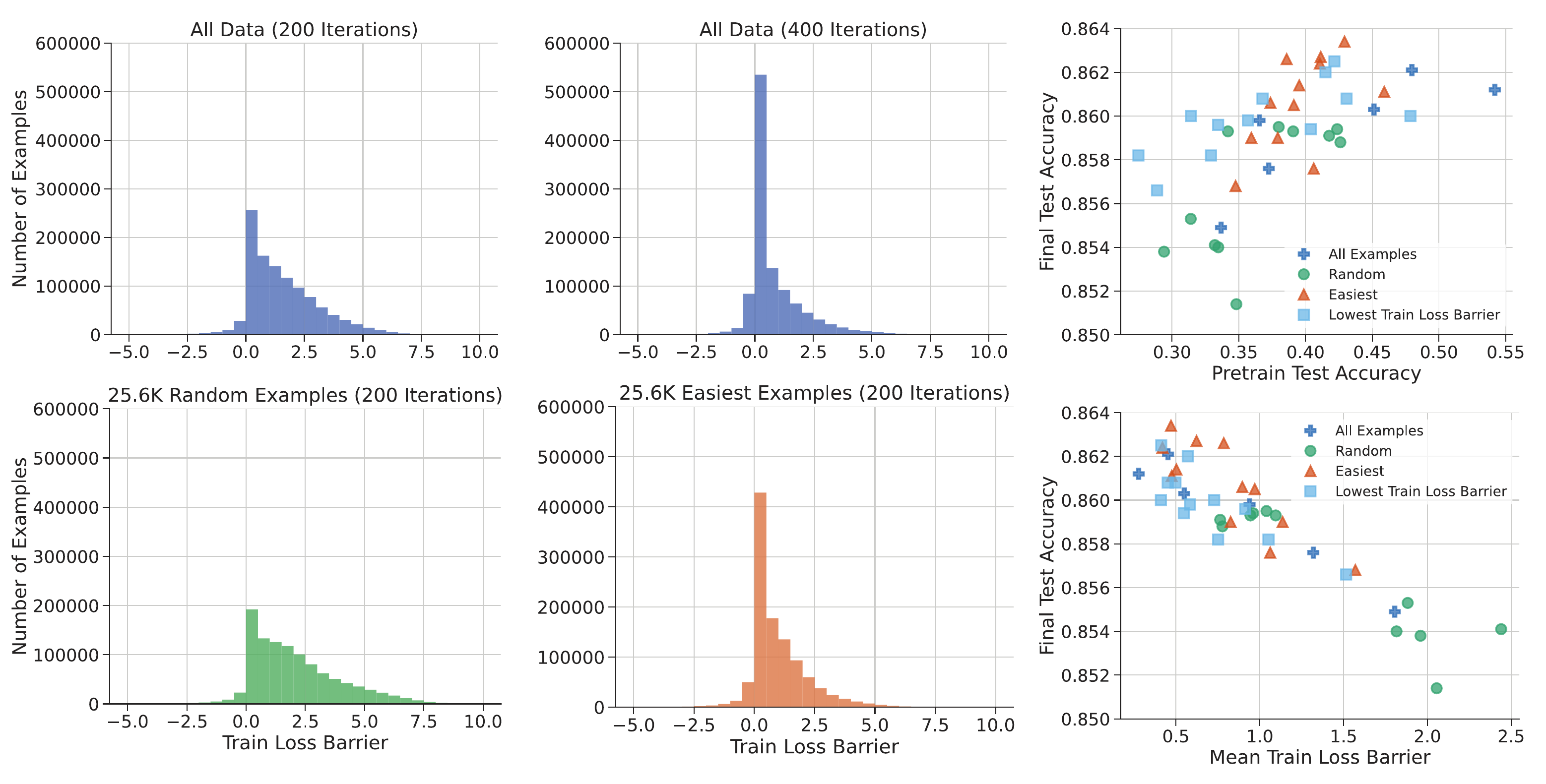}
	}
	\vspace{-1em}
	\caption{
	\textbf{Left and center columns:}
	Pre-training on all the data for $t^*$ (top right) vs. $t^*/2$ (top left) steps leads to a distribution shift to smaller per-examples train loss barriers between children runs of the dense network. When pre-trained for just $t^*/2$ iterations but with a random subset of data, the distribution is similar to pre-training on all data for $t^*/2$ steps. However, when pre-trained on a subset of easy examples for $t^*/2$ iterations, the distribution of train loss barriers displays a shift comparable to pre-training on all the data for $t^*$ iterations. 
	As seen in \cref{fig:Phase1AllSparsities} and the Appendix, pre-training with this random subset matches pre-training with all data for $t^*/2$ steps while pre-training with the easiest data for $t^*/2$ matches pre-training on all data for $t^*$ steps. This observation suggests that the per-example distribution of train loss barriers may be a useful signature for predicting the IMP performance of dense initializations.
	\textbf{Right column:} Scatter plots of final test accuracy vs. pre-train accuracy of the dense initialization (top) and the mean train loss barrier (bottom) across a variety of dataset pruning strategies for pre-training.
	Final test accuracy is at 8.6\% weights remaining (11 pruning rounds) for CIFAR-100 and 21\% weights remaining (7 pruning rounds) for CINIC-10.
	We observe that the mean train loss barrier is better correlated with the final test accuracy than the pre-train accuracy of the dense initialization.}
    \vspace{-1em}
	\label{fig:LMC-Results}
\end{figure*}

\section{Pre-training through the lens of Linear Mode Connectivity}
\label{sec:losslandscape}
\vspace{-0.5em}

\citet{frankle2019linear} observed a strong indicator of whether a \emph{sparse} sub-network generated by IMP would be able to match the accuracy of the dense network is whether or not that sparse sub-network train to the same linearly connected mode (i.e. the train loss barrier is close to 0, see \cref{fig:IMP-Illustration}).
Here we ask what properties of the \emph{dense} network at the rewind point might also be predictive of this property.
In the architectures + datasets we consider, the onset of linear mode connectivity (LMC) occurs in the dense network later than the first rewind time which produces a good initialization for IMP (i.e. $\bar{t} > \rt$), and thus, we consider two generalizations to the notion of LMC.  First, we look at the train loss barrier on a per-example basis meaning that instead of considering the full loss landscape we separately look at the performance of the linear interpolation on each example (the original notion of train loss barrier is obtained by averaging these values).  Second, we then look at the distribution of these per-example train loss barriers as a continuous measure of the state of the network rather than simply considering whether their average is 0.  We demonstrate in general that these measures of the loss landscape of the dense network correlate well with the IMP performance of the pre-trained initialization.

\vspace{-0.5em}
\paragraph{The distribution of per-example loss barriers.} In the left and center columns of \cref{fig:LMC-Results} we plot histograms of per-example loss barriers for various rewinding points and data subsets for CIFAR-100 (panel a) and CINIC-10 (panel b).
There is a clear shift in the empirical distribution of the per-example loss barriers --- not only the mean is shifting as previously observed, but also the mode. Pre-training on all data for $t^*$ iterations has the same effect on the distribution of the loss barriers as pre-training on the easy examples for $t^*/2$ iterations, two procedures which produce matching IMP initializations.

Given the connection between the error barrier and IMP sub-network performance,
it is natural to ask whether examples which achieve low loss barrier early in training with the full dataset are more important for producing a matching initialization than high loss barrier examples. 
We call this ranking of the data by the per-example loss barrier at a given iteration the LMC score, and the blue squares in the scatter plot are the result of training on a subset of the data determined by this score.
Our empirical findings suggest that using this score does not match the performance of using easy data (see \cref{app:lmcscores}).

\vspace{-0.5em}
\paragraph{The correlation between final test accuracy and mean train loss barrier.} The right column of \cref{fig:LMC-Results} shows a scatter plot of final test accuracy of the sparse trained network vs. two different properties of the dense initialization: test accuracy of this pre-train point (top) and the average train loss barrier from 3 pairs of children runs (bottom).
Here we observe the same result across the different training conditions: a correlation between the train loss barrier when spawning from $w_{\rt}$, and the final test accuracy of the sparse sub-networks.
The correlation is weaker for the pre-train test accuracy, suggesting that the loss landscape properties hold more import for determining the success of IMP.
Furthermore, we see that training on the easiest examples produces rewinding points with smaller train loss barriers than training on random subsets.

\section{The Role of Training Data in Learning Rate Warmup}

To this point, we have studied the early phase of training through the lens of the lottery ticket hypothesis.
However, the dynamics of the early phase of training have important implications beyond the sparse regime considered by lottery ticket research.
For example, learning rate warmup is a nearly ubiquitous part of training state-of-the-art models.
Warmup can be viewed as an accommodation for the idiosyncrasies of the earliest part of training and a pre-training strategy that prepares the network to train at the full learning rate.
In this section, we extend our previous experiments to dense training with learning rate warmup by ablating the training data during this phase to assess whether easy data alone suffices, similar to our observations in the \phaseone for IMP (\cref{sec:phase1}).

\begin{figure}[ht]
	\centering
	\includegraphics[width=\linewidth]{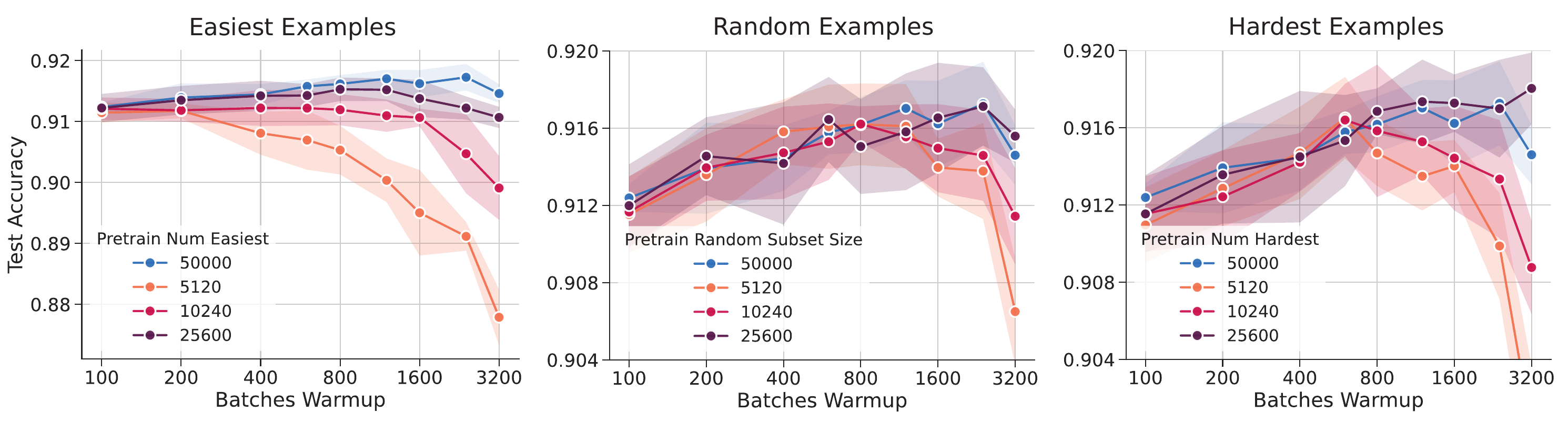}
	\vspace{-0.3cm}
	\caption{Training with different data subset sizes and pruning strategies for the learning rate warmup period for ResNet-20 on CIFAR-10.  In the large batch size and learning rate regime, learning rate warmup is essential for stabilizing training; we sweep over different learning rate warmup periods with a batch size of 1024.  In contrast to pre-training for IMP, training on easiest examples in this initial phase leads to worse performance than training with all data, random subsets, or the hardest data. We note that the learning rate warmup period is a much longer fraction of training than IMP pre-training which lasts for only a few hundred steps.}
	\vspace{-0.3cm}
	\label{fig:lrwarmup}
\end{figure}

\vspace{-0.5em}
\paragraph{Easy data does not reduce the amount of training required to stabilize the network via learning rate warmup.} The effects of performing learning rate warmup with easy, random, and hardest subsets of the data appear in \cref{fig:lrwarmup}.
Intriguingly, the most striking performance drop is observed when doing learning rate warmup with easy data, especially at longer pre-training times.
Pre-training on the random and hardest data produces the same performance for sufficiently large subset sizes.
The role played by data during IMP pre-training is thus different than that during learning rate warmup; the easy data does not add stability benefits during the later.
We hypothesize that this is because learning rate warmup happens over a much larger fraction of the total training time than IMP pre-training.

\section{Discussion}
\label{sec:discussion}

Recent empirical evidence has shown that deep neural network optimization proceeds in several distinct phases of training  \citep{Frankle2020The, fort2020deep}.
Understanding the role that data plays in these different phases can help us characterize what is being learned during them. Since data loading is often a bottleneck, this understanding also has the potential enable more efficient training schemes.  
In this work, we have considered two essential phases of early training: pre-training for IMP which enables sparse optimization and learning rate warmup which stabilizes network training.
Our experiments identify what data is sufficient for each of these procedures, and as they both occur early in training, these finding can be used to design more efficient data loaders for streaming datasets.
Furthermore, as IMP is a computationally expensive procedure, understanding how it works is essential for designing better algorithms with the same performance.
To this end, we identify loss landscape properties of the dense network initialization for IMP that are predictive of successful spare training.
Though this work does not provide an improved algorithm for obtaining sparse networks,
we believe our results provide essential guidance for researchers pursuing algorithms that perform pruning early in training (i.e. finding sparse masks without training to convergence).

\section*{Acknowledgements}
The experiments for this paper were funded by Google Cloud research credits.  S.G. thanks the James S. McDonnell and Simons Foundations, NTT Research, and an NSF CAREER Award for support while at Stanford.
This work was done in part while G.K.D. was visiting the Simons Institute for the Theory of Computing.
The authors would like to thank Daniel M. Roy for feedback on multiple drafts.

\bibliography{biblio}

\begin{thebibliography}{26}
\providecommand{\natexlab}[1]{#1}
\providecommand{\url}[1]{\texttt{#1}}
\expandafter\ifx\csname urlstyle\endcsname\relax
  \providecommand{\doi}[1]{doi: #1}\else
  \providecommand{\doi}{doi: \begingroup \urlstyle{rm}\Url}\fi

\bibitem[Baldock et~al.(2021)Baldock, Maennel, and Neyshabur]{baldock2021deep}
R.~Baldock, H.~Maennel, and B.~Neyshabur.
\newblock Deep learning through the lens of example difficulty.
\newblock \emph{Advances in Neural Information Processing Systems}, 34, 2021.

\bibitem[Chen et~al.(2020)Chen, Frankle, Chang, Liu, Zhang, Wang, and
  Carbin]{chen2020lottery}
T.~Chen, J.~Frankle, S.~Chang, S.~Liu, Y.~Zhang, Z.~Wang, and M.~Carbin.
\newblock The lottery ticket hypothesis for pre-trained bert networks.
\newblock \emph{Advances in Neural Information Processing Systems}, 2020.

\bibitem[Darlow et~al.(2018)Darlow, Crowley, Antoniou, and
  Storkey]{darlow2018cinic}
L.~N. Darlow, E.~J. Crowley, A.~Antoniou, and A.~J. Storkey.
\newblock Cinic-10 is not imagenet or cifar-10.
\newblock \emph{arXiv preprint arXiv:1810.03505}, 2018.

\bibitem[Draxler et~al.(2018{\natexlab{a}})Draxler, Veschgini, Salmhofer, and
  Hamprecht]{draxler2018essentially}
F.~Draxler, K.~Veschgini, M.~Salmhofer, and F.~Hamprecht.
\newblock Essentially no barriers in neural network energy landscape.
\newblock In \emph{International Conference on Machine Learning}, pages
  1309--1318. PMLR, 2018{\natexlab{a}}.

\bibitem[Draxler et~al.(2018{\natexlab{b}})Draxler, Veschgini, Salmhofer, and
  Hamprecht]{pmlr-v80-draxler18a}
F.~Draxler, K.~Veschgini, M.~Salmhofer, and F.~Hamprecht.
\newblock Essentially no barriers in neural network energy landscape.
\newblock In J.~Dy and A.~Krause, editors, \emph{Proceedings of the 35th
  International Conference on Machine Learning}, volume~80 of \emph{Proceedings
  of Machine Learning Research}, pages 1309--1318. PMLR, 10--15 Jul
  2018{\natexlab{b}}.
\newblock URL \url{https://proceedings.mlr.press/v80/draxler18a.html}.

\bibitem[Entezari et~al.(2021)Entezari, Sedghi, Saukh, and
  Neyshabur]{entezari2021role}
R.~Entezari, H.~Sedghi, O.~Saukh, and B.~Neyshabur.
\newblock The role of permutation invariance in linear mode connectivity of
  neural networks.
\newblock \emph{arXiv preprint arXiv:2110.06296}, 2021.

\bibitem[Evci et~al.(2020)Evci, Gale, Menick, Castro, and
  Elsen]{evci2020rigging}
U.~Evci, T.~Gale, J.~Menick, P.~S. Castro, and E.~Elsen.
\newblock Rigging the lottery: Making all tickets winners.
\newblock In \emph{International Conference on Machine Learning}, pages
  2943--2952. PMLR, 2020.

\bibitem[Fort et~al.(2020)Fort, Dziugaite, Paul, Kharaghani, Roy, and
  Ganguli]{fort2020deep}
S.~Fort, G.~K. Dziugaite, M.~Paul, S.~Kharaghani, D.~M. Roy, and S.~Ganguli.
\newblock Deep learning versus kernel learning: an empirical study of loss
  landscape geometry and the time evolution of the neural tangent kernel.
\newblock \emph{arXiv preprint arXiv:2010.15110}, 2020.

\bibitem[Frankle and Carbin(2018)]{frankle2018lottery}
J.~Frankle and M.~Carbin.
\newblock The lottery ticket hypothesis: Finding sparse, trainable neural
  networks.
\newblock \emph{arXiv preprint arXiv:1803.03635}, 2018.

\bibitem[Frankle et~al.(2020{\natexlab{a}})Frankle, Dziugaite, Roy, and
  Carbin]{frankle2019linear}
J.~Frankle, G.~K. Dziugaite, D.~M. Roy, and M.~Carbin.
\newblock Linear mode connectivity and the lottery ticket hypothesis.
\newblock In \emph{Proc. Int. Conf. Machine Learning (ICML)},
  2020{\natexlab{a}}.

\bibitem[Frankle et~al.(2020{\natexlab{b}})Frankle, Schwab, and
  Morcos]{Frankle2020The}
J.~Frankle, D.~J. Schwab, and A.~S. Morcos.
\newblock The early phase of neural network training.
\newblock In \emph{International Conference on Learning Representations},
  2020{\natexlab{b}}.
\newblock URL \url{https://openreview.net/forum?id=Hkl1iRNFwS}.

\bibitem[Garipov et~al.(2018)Garipov, Izmailov, Podoprikhin, Vetrov, and
  Wilson]{garipov2018loss}
T.~Garipov, P.~Izmailov, D.~Podoprikhin, D.~P. Vetrov, and A.~G. Wilson.
\newblock Loss surfaces, mode connectivity, and fast ensembling of dnns.
\newblock \emph{Advances in Neural Information Processing Systems}, 31, 2018.

\bibitem[Gilmer et~al.(2021)Gilmer, Ghorbani, Garg, Kudugunta, Neyshabur,
  Cardoze, Dahl, Nado, and Firat]{gilmer2021loss}
J.~Gilmer, B.~Ghorbani, A.~Garg, S.~Kudugunta, B.~Neyshabur, D.~Cardoze, G.~E.
  Dahl, Z.~Nado, and O.~Firat.
\newblock A loss curvature perspective on training instabilities of deep
  learning models.
\newblock In \emph{International Conference on Learning Representations}, 2021.

\bibitem[Han et~al.(2015)Han, Pool, Tran, and Dally]{han2015learning}
S.~Han, J.~Pool, J.~Tran, and W.~J. Dally.
\newblock Learning both weights and connections for efficient neural network.
\newblock In \emph{Advances in Neural Information Processing Systems}, 2015.

\bibitem[He et~al.(2016)He, Zhang, Ren, and Sun]{he2016deep}
K.~He, X.~Zhang, S.~Ren, and J.~Sun.
\newblock Deep residual learning for image recognition.
\newblock In \emph{Proceedings of the IEEE conference on computer vision and
  pattern recognition}, pages 770--778, 2016.

\bibitem[Janowsky(1989)]{janowsky1989pruning}
S.~A. Janowsky.
\newblock Pruning versus clipping in neural networks.
\newblock \emph{Physical Review A}, 39\penalty0 (12):\penalty0 6600, 1989.

\bibitem[Kalibhat et~al.(2020)Kalibhat, Balaji, and Feizi]{kalibhat2020winning}
N.~M. Kalibhat, Y.~Balaji, and S.~Feizi.
\newblock Winning lottery tickets in deep generative models.
\newblock \emph{arXiv preprint arXiv:2010.02350}, 2020.

\bibitem[Krizhevsky et~al.(2014)Krizhevsky, Nair, and
  Hinton]{krizhevsky2014cifar}
A.~Krizhevsky, V.~Nair, and G.~Hinton.
\newblock The cifar-10 dataset.
\newblock \emph{online: http://www. cs. toronto. edu/kriz/cifar. html},
  55:\penalty0 5, 2014.

\bibitem[Kuditipudi et~al.(2019)Kuditipudi, Wang, Lee, Zhang, Li, Hu, Ge, and
  Arora]{kuditipudi2019explaining}
R.~Kuditipudi, X.~Wang, H.~Lee, Y.~Zhang, Z.~Li, W.~Hu, R.~Ge, and S.~Arora.
\newblock Explaining landscape connectivity of low-cost solutions for
  multilayer nets.
\newblock \emph{Advances in Neural Information Processing Systems}, 32, 2019.

\bibitem[LeCun et~al.(1990)LeCun, Denker, and Solla]{lecun1990optimal}
Y.~LeCun, J.~S. Denker, and S.~A. Solla.
\newblock Optimal brain damage.
\newblock In \emph{Advances in Neural Information Processing Systems}, pages
  598--605, 1990.

\bibitem[Nagarajan and Kolter(2019)]{nagarajan2019uniform}
V.~Nagarajan and J.~Z. Kolter.
\newblock Uniform convergence may be unable to explain generalization in deep
  learning.
\newblock \emph{Advances in Neural Information Processing Systems}, 32, 2019.

\bibitem[Paul et~al.(2021)Paul, Ganguli, and Dziugaite]{paul2021deep}
M.~Paul, S.~Ganguli, and G.~K. Dziugaite.
\newblock Deep learning on a data diet: Finding important examples early in
  training.
\newblock \emph{arXiv preprint arXiv:2107.07075}, 2021.

\bibitem[Vischer et~al.(2021)Vischer, Lange, and Sprekeler]{vischer2021lottery}
M.~A. Vischer, R.~T. Lange, and H.~Sprekeler.
\newblock On lottery tickets and minimal task representations in deep
  reinforcement learning.
\newblock \emph{arXiv preprint arXiv:2105.01648}, 2021.

\bibitem[Yu et~al.(2020)Yu, Edunov, Tian, and Morcos]{yu2020playing}
H.~Yu, S.~Edunov, Y.~Tian, and A.~S. Morcos.
\newblock Playing the lottery with rewards and multiple languages: lottery
  tickets in rl and nlp.
\newblock In \emph{International Conference on Learning Representations}, 2020.
\newblock URL \url{https://openreview.net/forum?id=S1xnXRVFwH}.

\bibitem[Zhang et~al.(2016)Zhang, Bengio, Hardt, Recht, and
  Vinyals]{Zhang2016UnderstandingDL}
C.~Zhang, S.~Bengio, M.~Hardt, B.~Recht, and O.~Vinyals.
\newblock Understanding deep learning requires rethinking generalization.
\newblock \emph{CoRR}, abs/1611.03530, 2016.

\bibitem[Zhu and Gupta(2017)]{zhu2017prune}
M.~Zhu and S.~Gupta.
\newblock To prune, or not to prune: exploring the efficacy of pruning for
  model compression.
\newblock \emph{arXiv preprint arXiv:1710.01878}, 2017.

\end{thebibliography}
 
\newpage

\appendix
\onecolumn

\section{Experimental Details}
\label{app:expdetails}

\paragraph{Code.}  The code used to run the experiments is available at:
\url{https://github.com/mansheej/lth_diet}

\paragraph{Datasets.} We used CIFAR-10, CIFAR-100 \citep{krizhevsky2014cifar}, and CINIC-10 \citep{darlow2018cinic} in our experiments.  For CINIC-10, we combine the training and validation sets into a single training set with 180,000 images. The standard test set of 90,000 images is used for testing. Each dataset is normalized by its per channel mean and standard deviation over the training set. All datasets get the same data augmentation: pad by 4 pixels on all sides, random crop to 32$\times$32 pixels, and left-right flip image with probability half.

\paragraph{Models.}  In these experiments we use ResNet-20, ResNet-32, ResNet-56 \citep{he2016deep}. These are the low-resolution CIFAR variants of ResNets from the original paper. The variants of the network used are specified in the figures.

\paragraph{Randomized Labels.}  The labels were randomized during the pre-training phase \textit{only} by first selecting 10\%/50\% of the training uniformly at random and then drawing a new label uniformly from the 10 classes of CIFAR-10.  Note that because this procedure can result in an example being reassinged the correct label, on average only 9\%/45\% of the labels are corrupted by this procedure.  The dataset is corrupted once and then resused across all subset sizes and replicates.

The EL2N scores used to determine the subset of easiest data are computed by training on the corrupted dataset.  As seen in Figure 5 of \citep{baldock2021deep}, the network typically learns the correct label early in training for corrupted data points. As a result, the corrupted examples will be ranked as difficult by the EL2N scores as verified in \citep{paul2021deep}.  Thus, the noisy labels are filtered out by pre-training on the examples with the lowest EL2N scores.

\paragraph{Linear mode connectivity procedure.} To determine if two children of a given network are linearly mode connected, we perform the following procedure: 
\begin{enumerate}[leftmargin=2em]
    \item Spawn two children initialized with the current weights of the given network.
    \item Independently train the two networks with different SGD noise realized through different data orders.
    \item Calculate the mid-point in weight space between the two networks.
    \item Calculate the training loss for the resulting network at the mid-point.
    \item Calculate the training loss barrier---the loss calculated at the mid-point minus the mean of the losses calculated at the endpoints.
\end{enumerate}
Train loss barriers are calculated both on a per-example basis and as averages over the whole dataset. The per-example training loss barrier is just the training loss barrier evaluated on that example. In practice, we train three children and calculate the train loss barrier between each pair and average.
Following \citep{frankle2019linear}, we say that two networks are linearly mode connected if the train loss barrier between them is less than 2\%.

\paragraph{EL2N score computation} To calculate EL2N scores for a dataset, we follow the process outlined in \citep{paul2021deep}. In particular, we do the following:
\begin{enumerate}[leftmargin=2em]
    \item Independently train $K=10$ networks from different random initializations for $t$ iterations. 
    \item For each example and each network, we calculate the L2 norm of the error vector defined as $\|p(\V{x}) - \V{y}\|_2$ where $\V{y}$ is the one-hot encoding of the label, and $p(\V{x})$ are the softmax outputs of the network evaluated on example $\V{x}$.
    \item For each example, the EL2N score is the average of the error vector L2 norm across the $K$ networks.
\end{enumerate}
To calculate these scores, we use ResNet-20 and $t=7800$ iterations for CIFAR-10, ResNet-32 and $t=7800$ iterations for CIFAR-100, and ResNet-56 and $t=8000$ iterations for CINIC-10.
\label{app:EL2N_calc}

\paragraph{Hyperparameters.} Networks were trained with stochastic gradient descent (SGD). $t^*$ was chosen for each dataset such that $t_r = t^*$ produces sparse networks that can be trained to the same accuracy as the dense network to atleast 16.8\% sparsity (8 pruning levels). The pre-training learning rate was chosen based on which of the set $\{0.1, 0.2, 0.4\}$ produced the best performance at $t^*$. The full hyperparameters are provided in \cref{tab:hyperparameters}.

\begin{table*}[!ht]
  \centering
  \vspace{-0.25cm}
  \caption{Hyperparameters Used for Experiments.}
  \vspace{0.2cm}
  \begin{tabular}{c||c|c|c}
         & \bf CIFAR-10 & \bf CIFAR-100 & \bf CINIC-10 \\
        \hline
        \hline
        ResNet Variant & ResNet-20 & ResNet-32 & ResNet-56 \\
        Batch Size & 128 & 128 & 256 \\
        Pre-training Learning Rate & 0.4 & 0.4 & 0.1 \\
        Learning Rate & 0.1 &  0.1 &  0.1 \\
        Momentum & 0.9 & 0.9 & 0.9 \\
        Weight Decay & 0.0001 & 0.0001 & 0.0001 \\
        Learning Rate Decay Factor & 0.1 & 0.1 & 0.1 \\
        Learning Rate Decay Milestones & 31200, 46800 & 31200, 46800 &  15625, 23440\\
        Total Training Iterations & 62400 & 62400 & 31250 \\
        IMP Weight Pruning Fraction & 20\% & 20\% & 20\%
    \end{tabular}
  \label{tab:hyperparameters}
\end{table*}

All results are reported as the mean and standard deviation of 4 replicates. The only exception is \cref{fig:LMC-Results} scatter plots  where each point corresponds to 1 replicate.

\paragraph{Learning rate warmup experimental design.} 
For the learning rate warm-up experiments, we train ResNet-20 on CIFAR-10 with batch size = 1024. For the optimizer, we use SGD with learning rate = 3.2, momentum = 0.9 and weight decay = 0.0001. There is an initial linear learning rate warm-up for $t$ batches. The network is then trained for a total of 160 epochs with a LR drop by a factor of 10 at 80 and 120 epochs. In \cref{fig:lrwarmup}, we sweep the learning rate warmup period $t$. For each hyperparameter configuration, we report the mean and standard deviation of 8 replicates. In addition to all the training data, we perform experiments with different subsets of the training data with sizes 5120, 10240 and 25600. In these cases, the network was trained on the data subset during the warm-up period only and on the full dataset afterwards. For the easiest (hardest) examples, we choose examples with the lowest (highest) EL2N scores which were calculated as described above.

\paragraph{Compute Resources.}
The experiments were performed on virtual Google Cloud instances configured with 4 NVIDIA Tesla A100 GPUs.  Each experiment replicate was run on a single A100 GPU.  The approximate compute time for a full run of IMP was 8 hours for CIFAR-10, 12 hours for CIFAR-100, and 10 hours for CINIC-10.

\paragraph{Ethical and societal consequences.} Being empirically-driven work, the experiments performed consumed considerable energy.  However, we view our work as a step towards the long-term goal of improving the efficiency of training neural network which will ideally lead to an increase in the democratization and sustainability of AI.  Towards this aim, our paper provides insight into the role of data in early training and a better understanding of which properties of pre-trained initializations enable sparse optimization.  We hope this will inspire better algorithms for training sparse networks.

\section{Other hypothesis for the role of easy data}
\label{app:otherhypothesis}

Pre-training on easy data (lowest EL2N scores) allows us to reduce $\rt$ without hurting test accuracy of the sparse networks found by IMP. One possible reason for this is that networks pre-trained on just the easy data achieve better test accuracies at step $\rt$. But in \cref{fig:LMC-Results}, we find that the accuracy of the dense network at the pre-trained initialization does not completely explain the improved performance. Here we explore two additional hypotheses for what role the easy data plays during the pre-training phase.  

 \subsection{Changes in the gradient size.}
 \label{app:gradsize}

Examples with lower EL2N scores have smaller gradient norms at the point at which they are computed as observed by \citep{paul2021deep}. 
We investigate whether the gradient norm is also smaller when training on the easiest examples during the pre-training phase of IMP, which would reproduce the effects of gradient clipping which is often beneficial early in training. 
We compare the distribution of batch gradient norms seen in the first 200 steps while training on easy data versus a random subset of data. Note that since we are only training for 200 steps with a batch size of 128, training on all the data is equivalent to training on random data. We run the following experiment:

\begin{enumerate}[leftmargin=2em]
    \item Train a ResNet-20 on CIFAR-10 for 200 iterations on a random subset of 25600 examples. At every iteration record the gradient norm of the batch.
    \item Repeat the above process but training on a subset of 25600 examples with the smallest EL2N scores.
    \item Compare the distribution of gradient norms
\end{enumerate}

\begin{figure}[!ht]
	\centering
	\includegraphics[width=0.45\linewidth]{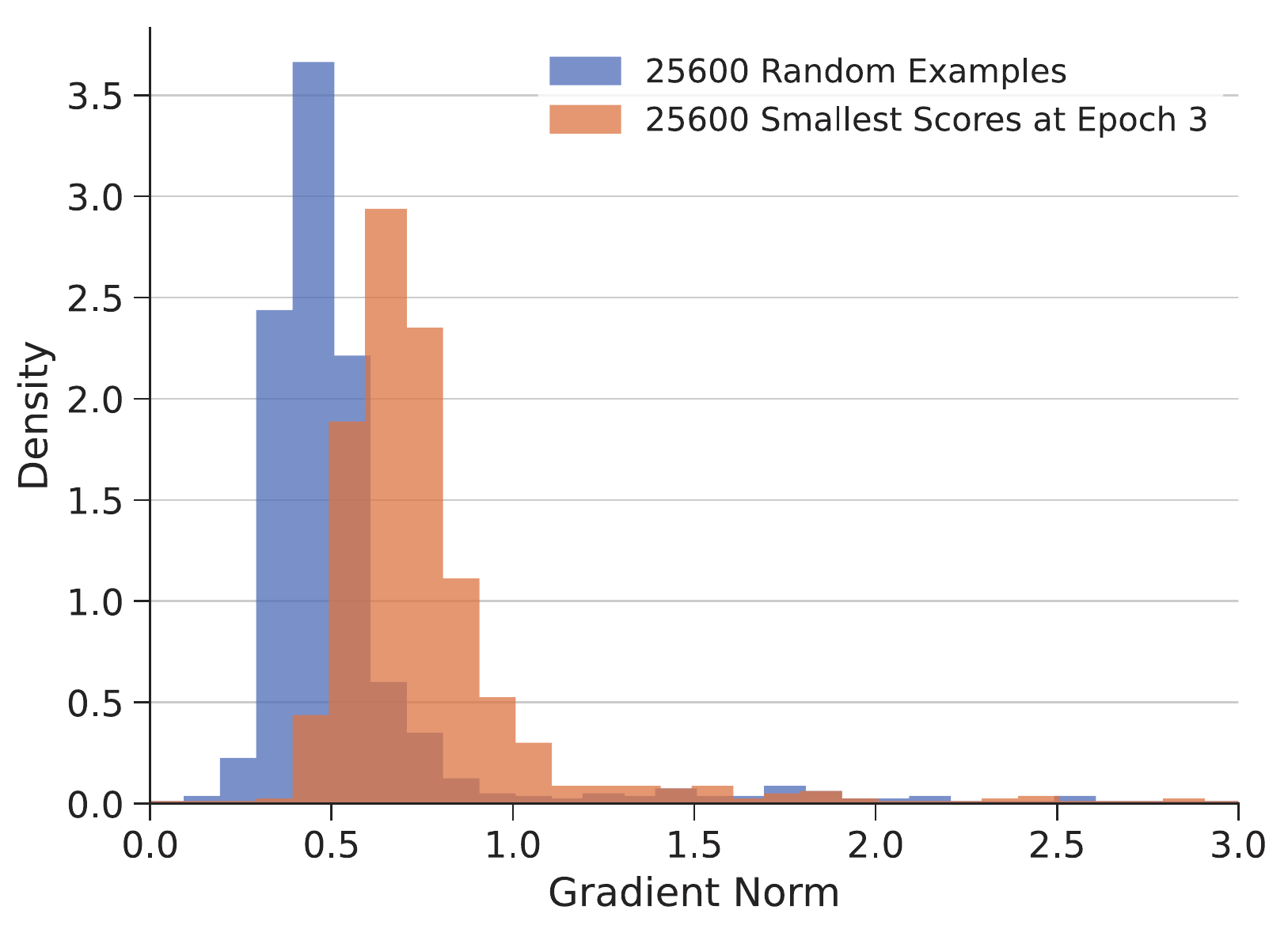}
	\vspace{-0.4cm}
	\caption{The histogram of minibatch gradient norms during the first $\rt=200$ iterations of training (CIFAR-10, ResNet-20), when the minibatches are sampled from a random subset of 25600 examples compared to sampled from a subset of the 25600 easiest examples.}
	\label{fig:Grad-Size}
	\vspace{-0.4cm}
\end{figure}

\cref{fig:Grad-Size} shows the results. We find that, when training on easy examples, the batch gradient norms are typically larger. Our hypothesis is that on easy examples, the gradients are more aligned, leading to larger average gradients.
We thus reject the hypothesis that pre-training on the easy subset of the data effectively performs gradient clipping.

\subsection{Change in distribution of per-example error}

\begin{figure}[!ht]
	\centering
	\includegraphics[width=\linewidth]{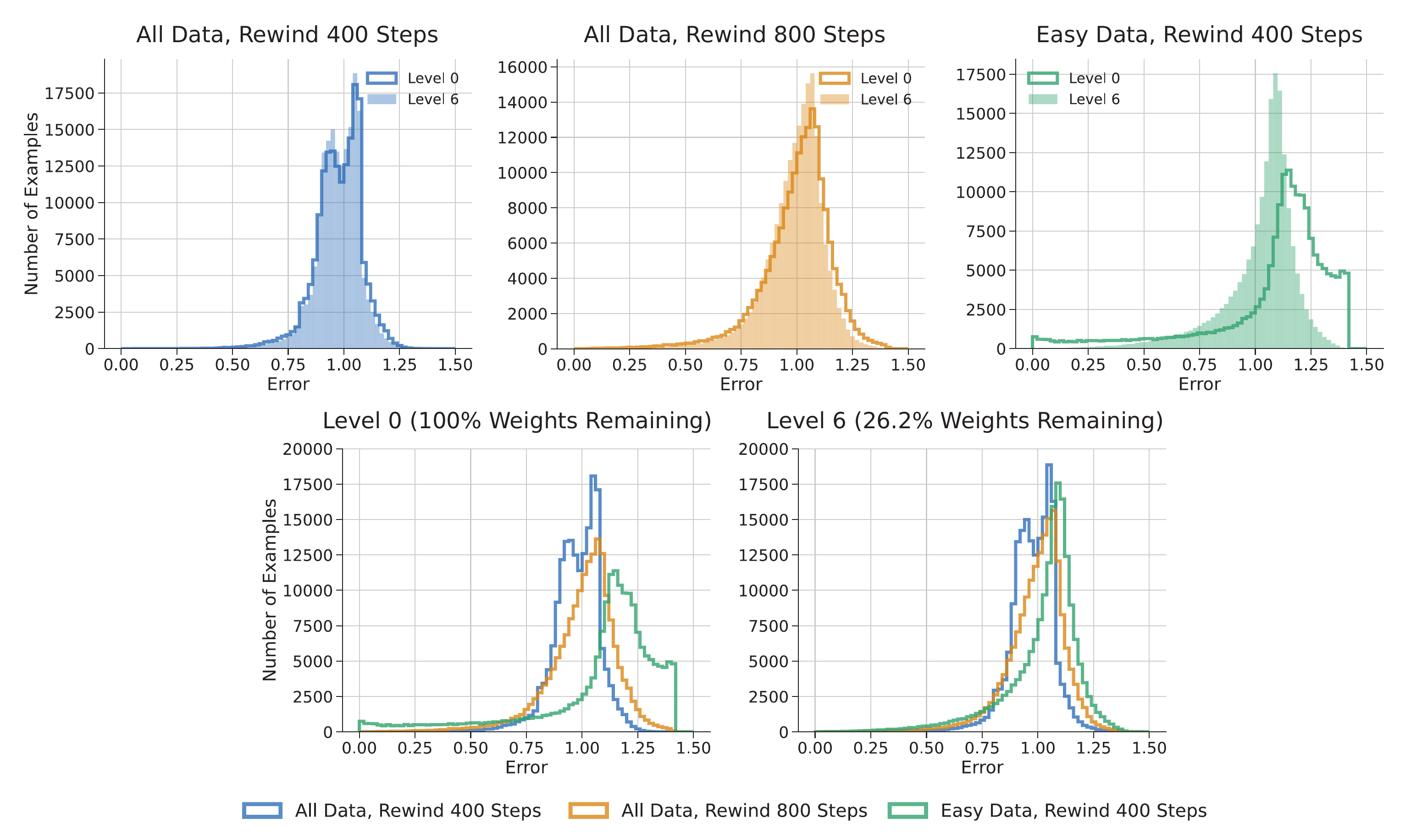}
	\vspace{-0.4cm}
	\caption{The distribution of per-example error at the pre-training point $t_r$ for several different pre-training procedures before and after sparse projection. These errors are calculated for a single ResNet-56 network on CINIC-10 (i.e. not averaged across replicates).  The top row shows the distributions grouped by pre-training procedure while the bottom row shows all error distributions in the dense networks (left) and sparse networks (right).  
	}
	\label{fig:Error-Distribution}
	\vspace{-0.4cm}
\end{figure}

For a ResNet-56 trained on CINIC-10, we considered whether different pre-training procedures resulted in diffferent distributions of the per-example error early in training.  We made the following observations in the results shown in \cref{fig:Error-Distribution}

\begin{enumerate}[leftmargin=2em]
    \item Prior to the sparse projection, there is a distinctive distribution of error after training on easy data.  A split occurs in the data where a portion of the examples shifts towards zero error while the remainder increases in error as compared to training on all data (level 0 green curve vs. level 0 blue curve in \cref{fig:Error-Distribution}).
    \item When comparing the dense vs. sparse networks pre-trained on all data, the distribution looks the same before and after projection (level 0 vs. level 6 blue and orange curves).  However, for pre-training on the easy data a significant change happens after the projection (level 0 vs. level 6 green curves).
    \item After the projection, the error distributions from training on all data for 800 iterations and easy data for 400 iterations are similar (level 6 orange and green curve) while the training on all data for 400 iterations is different (level 6 blue curve).
\end{enumerate}

These observations bear similarity to the per-example train loss barrier results considered in the main text (\cref{fig:LMC-Results}). Pre-training on easy data for 400 iterations produces results that have the same signatures as pre-training on all data for 800 iterations (and differ from training on all data for 400 iterations).  However, the per-example train loss barrier provides a clearer explanation and thus we think is the more fundamental quantity.

\section{Linear Mode Connetivity: Additional Experiments}
\label{app:lmcscores}

Here we present additional work on understanding the connection between linear mode connectivity of a dense initialization and its performance as a pre-trained initialization for IMP.  We introduce a new score for ranking examples in terms of the per-example train loss barrier early in training which we call the LMC score.  We then compare this score to EL2N and test the IMP performance of pre-training on the subset of examples with low LMC scores.

\subsection{Background on (linear) mode connectivity.}
The deep learning phenomena of \emph{mode connectivity} via simple paths has been uncovered by
\citet{pmlr-v80-draxler18a,garipov2018loss}.
In particular, it was shown that SGD-trained deep neural network modes are connected via simple nonlinear paths, along which the training and test loss is approximately constant.
Since then this phenomena has also been studies theoretically \citep{kuditipudi2019explaining}.
A related mode property has been described in the literature on generalization in deep learning: \citet{nagarajan2019uniform} show that for small networks trained on MNIST there exist \emph{linearly} connected modes.
Surprisingly, a slightly modified version of this phenomena is also true for large scale vision networks:
\citet{frankle2019linear} show that coupling the first training epochs restricts SGD to converge to the same linearly connected mode.
More recently, \citet{entezari2021role} hypothesize that linearly connected modes are equivalent up to symmetries. Their empirical observations suggest that there is overlap between the functions parameterized by different modes.

\subsection{LMC Scores}

Our observation that successful pre-train initializations are correlated with a shift to smaller per-example train loss barriers (left and center columns of \cref{fig:LMC-Results}) naturally raises the question: are examples that become linearly mode connected first also the ones which are important for this phase of training?
To this end, we introduced the notion of a per-example LMC score which allows us to rank examples in this manner:

\begin{definition}
The \emph{LMC score} of an example is the loss barrier computed on a dense network at rewinding iteration $t$ averaged over $K$ instances.
\end{definition}

\begin{figure*}[!ht]
    \subfigure[CIFAR-100, ResNet-32, Batch Size 128.]{
	\centering
	\includegraphics[width=\linewidth]{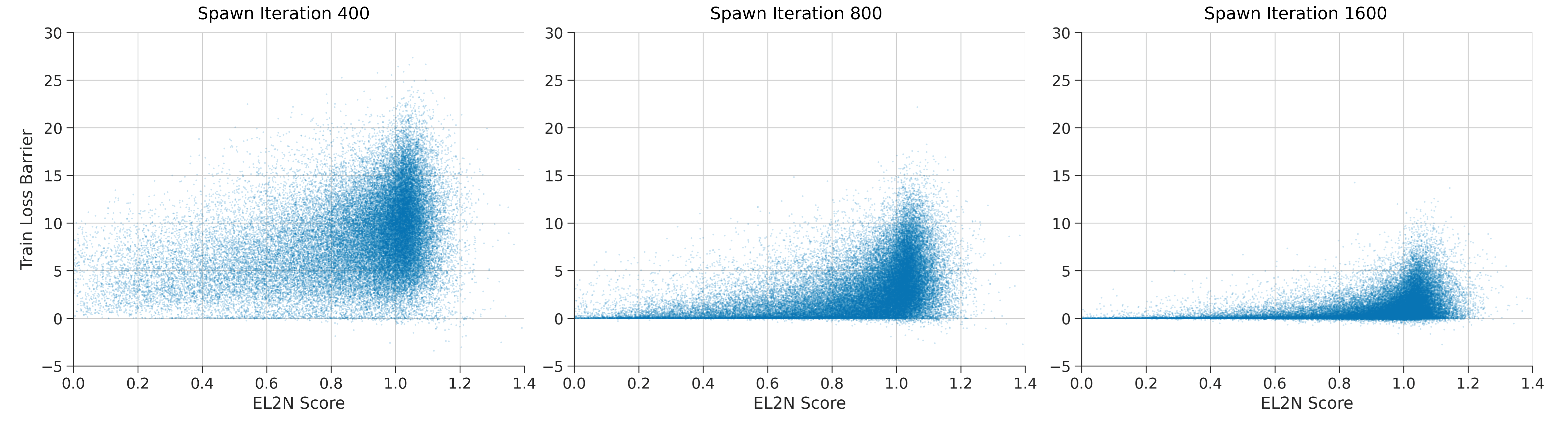}
	}
	\subfigure[CINIC-10, ResNet-56, Batch Size 256.]{
	\centering
	\includegraphics[width=\linewidth]{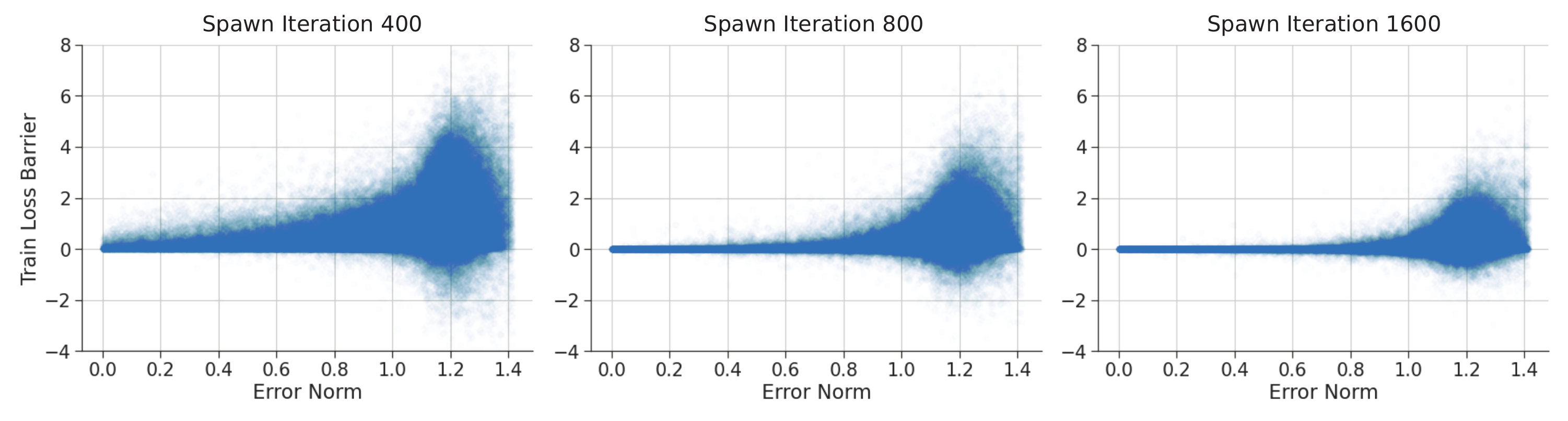}
	}
	\caption{Scatter plot of the relationship between the EL2N score computed at 1600 iterations and the Linear Mode Connectivity (LMC) score computed at iteration 400, 800, and 1600 iterations.}
	\label{fig:EL2N-LMC-Correlation}
\end{figure*}

\begin{figure*}[!ht]
	\subfigure[CIFAR-100, ResNet-32, $t^*/2 = 400$  LMC scores computed after 1600 iterations (batch size 128).]{
	\centering
	\includegraphics[width=\linewidth]{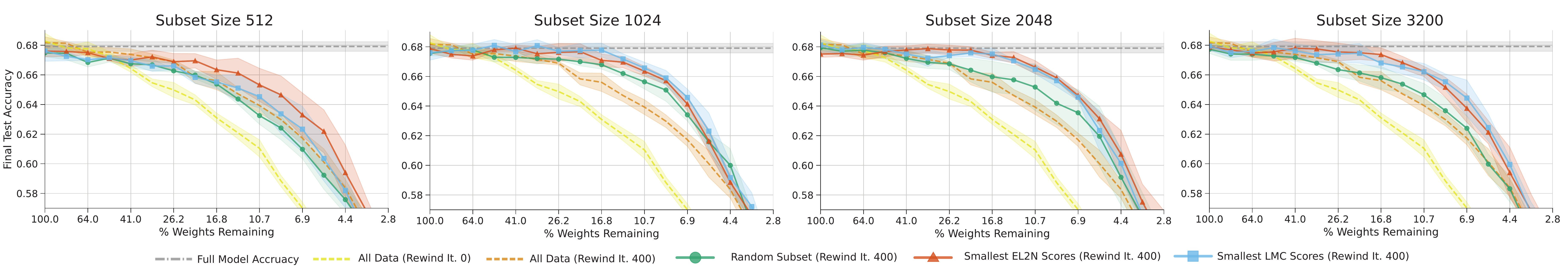}
	}
	\subfigure[CINIC-10, ResNet-56, $t^*/2 = 400$. LMC scores computed after 800 iterations (batch size 256).]{
	\centering
	\includegraphics[width=\linewidth]{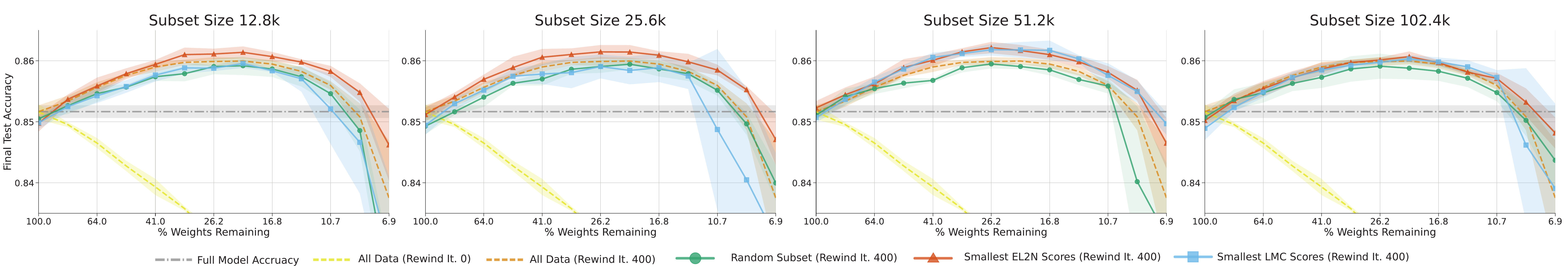}
	}
	\caption{Performance of IMP pre-training using data subsets determined by smallest LMC scores (solid light blue curve with squares).  This is compared against random subsets (solid green curve with circles) and easiest examples (solid red curve with triangles).  Pre-training is performed with learning rate 0.4 and batch size 128 for CIFAR-100 and learning rate 0.1 and batch size 256 for CINIC-10.  Several of these runs are included in the scatters plot in the right column of \cref{fig:LMC-Results} with the method labelled as ``Lowest Train Loss Barrier."}
	\label{fig:lmc-scores-full}
\end{figure*}

Under this definition, the train error barrier is the LMC scores averaged over the training data. If we compute the score after the onset of linear mode connectivity $\bar{t}$ it will not provide an informative ranking of the data as the score will be close to 0 for all examples; we compute the scores before $\bar{t}$. In our experiments, we computed the LMC score by averaging over three children runs from two replicates, so that each replicate provided three train loss barriers (one for each pair of children) and the total average was taken over six instances.

We visualize the correlation between EL2N scores and LMC scores. The scatter plots in Figure \cref{fig:EL2N-LMC-Correlation} show that low EL2N score examples tend to be the ones for which the LMC score goes down rapidly with spawning time, and is nearly-zero early in training. However, the opposite is not true, and high EL2N scores have a wide range of per-example loss barriers.

\subsection{Pre-training on Low LMC Scores}

\cref{fig:lmc-scores-full} shows that rewinding points obtained by training on low LMC score examples dominate random example pre-training, but are (weakly) dominated by training on low EL2N score examples.
We thus conclude that low EL2N examples are more representative of the necessary and sufficient subset of examples needed to find a good rewinding point.

\section{Full Results}
\label{sec:FullResults}

\begin{figure*}[!ht]
	\subfigure[CIFAR-100, ResNet-32, $t^*=800$, $t^*/2 = 400$.]{
	\centering
	\includegraphics[width=\linewidth]{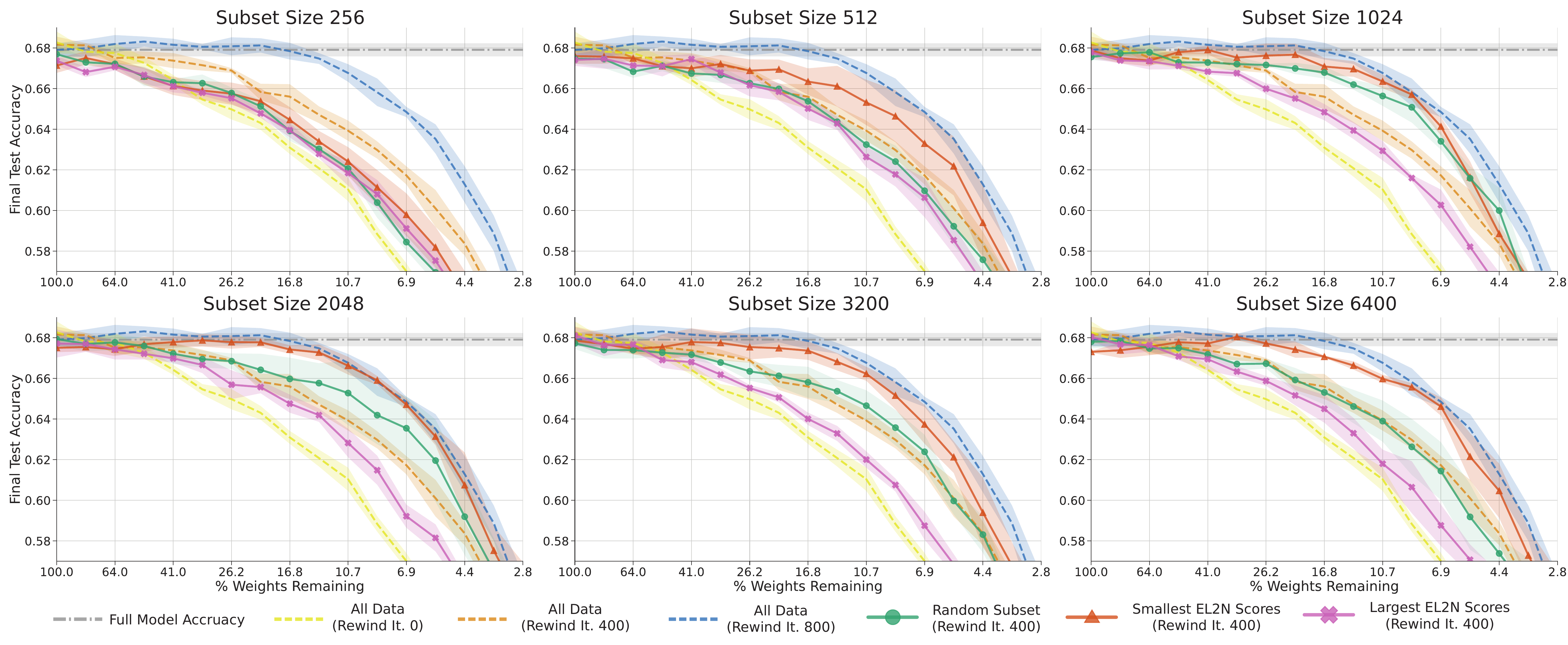}
	}
	\subfigure[CIFAR-10, ResNet-20, $t^*=400$, $t^*/2 = 200$.]{
	\centering
	\includegraphics[width=\linewidth]{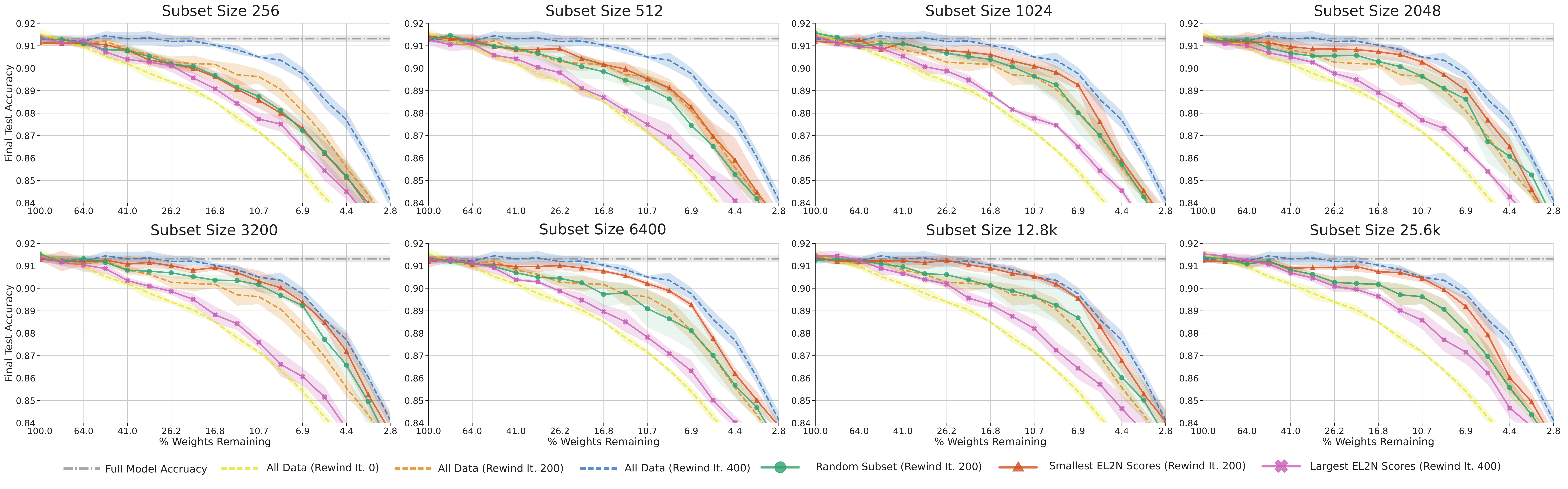}
	}
	\caption{Extended results for \cref{fig:Phase1AllSparsities} and \cref{fig:Phase1Subset} across all subset sizes considered on CIFAR-10 and CIFAR-100.  For each dataset, pretraining was performed with all data for $t^*/2$ steps (dashed orange curve) and $t^*$ steps (dashed blue curve).  Performing IMP as an initialization is included as a baseline (dashed yellow curve).  IMP pre-training was then performed with three different data subsets for $t^*/2$ steps: random examples (solid green curve with circles), easiest examples (solid red curve with triangles), and hardest examples (solid pink curve with crosses).  For both CIFAR-100 and CIFAR-10, a learning rate of 0.4 and batch size 128 was used during the pre-training period.}
	\label{fig:cifar-full}
\end{figure*}

\begin{figure*}[!ht]
	\subfigure[CINIC-10, ResNet-56, $t^* = 400$, $t^*/2 = 200$.]{
	\centering
	\includegraphics[width=\linewidth]{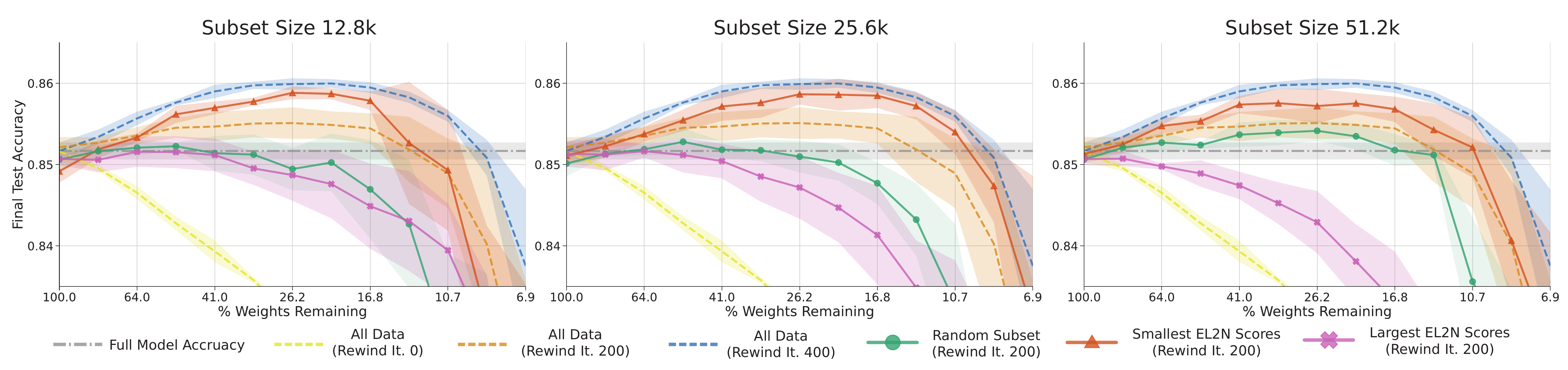}
	}
	\subfigure[CINIC-10, ResNet-56, $t^*=800$, $t^*/2 = 400$.]{
	\centering
	\includegraphics[width=\linewidth]{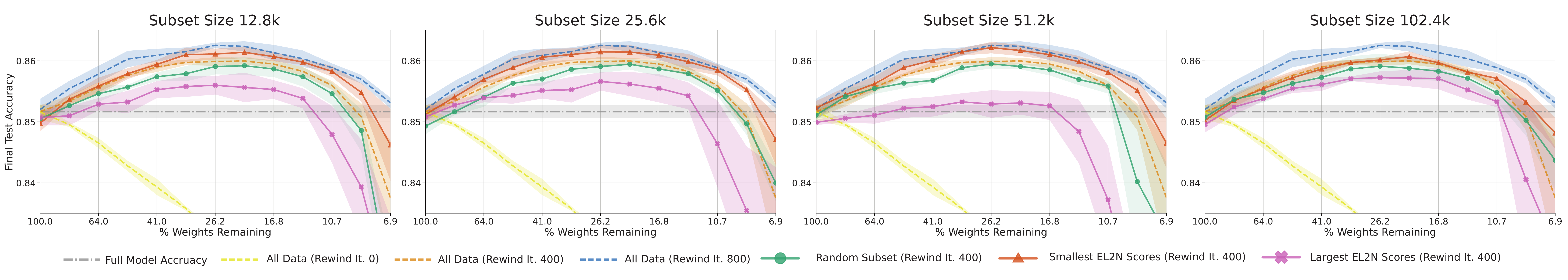}
	}
	\caption{Extended results for \cref{fig:Phase1AllSparsities} and \cref{fig:Phase1Subset} across all subset sizes considered on CINIC-10.  For each dataset, pretraining was performed with all data for $t^*/2$ steps (dashed orange curve) and $t^*$ steps (dashed blue curve).  Performing IMP as an initialization is included as a baseline (dashed yellow curve).  IMP pre-training was then performed with three different data subsets for $t^*/2$ steps: random examples (solid green curve with circles), easiest examples (solid red curve with triangles), and hardest examples (solid pink curve with crosses).  For CINIC-10, a learning rate of 0.1 and batch size 256 was used during the pre-training period.}
	\label{fig:cinic-full}
\end{figure*}

\begin{figure*}[!ht]
	\includegraphics[width=\linewidth]{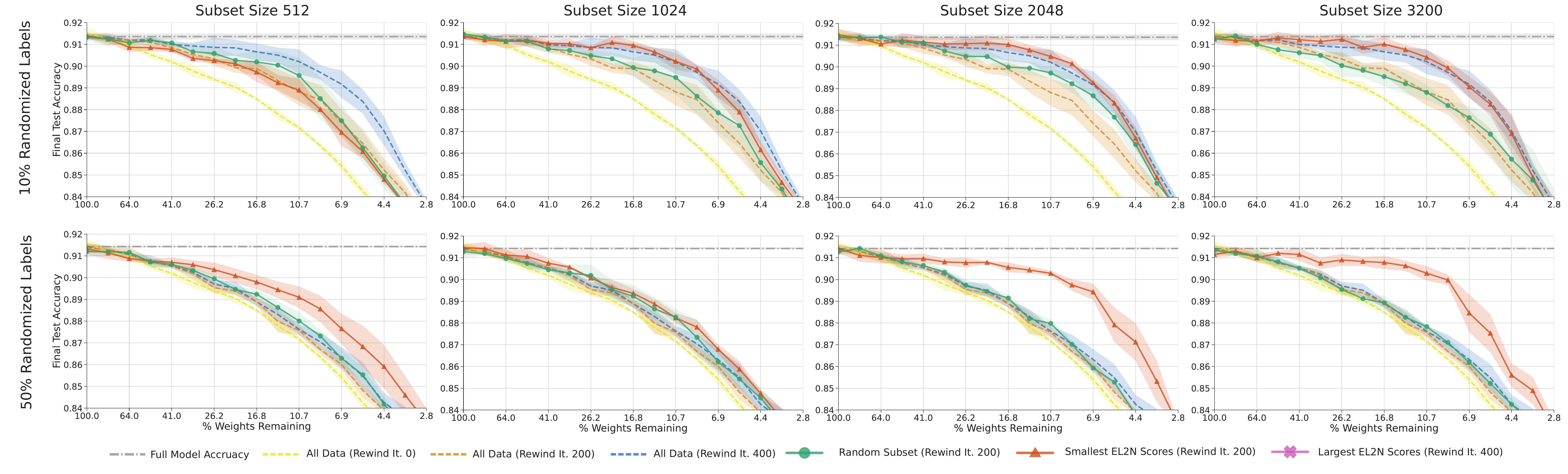}
	\caption{Extended results for \cref{fig:randomdatapretraining}.  In the top row, pre-training is performed on the dataset with 10\% randomly corrupted labels; in the bottom row, pre-training is performed with 50\% randomly corrupted labels.  The corruption is performed once and then is held the same across subset sizes and replicates.  Here $t^* = 400$, and the random and easy data subsets are trained for $t^*/2 = 200$ steps.  Pre-training was performed with a learning rate of 0.4 and batch size of 128.}
	\label{fig:rand-labels-full}
\end{figure*}

Here we present the full set of experiments performed for the results in the main text.  
\Cref{fig:cifar-full} and \cref{fig:cinic-full} show the results for performing IMP pre-training with different data subsets across a range of different subset sizes.  \Cref{fig:rand-labels-full} shows the result for performing IMP pre-training with the dataset corrupted by randomized label noise (the original dataset is then used for the mask search and sparse training phase).

\end{document}